\documentclass[preprint,5p]{elsarticle}

\usepackage{prletters}
\usepackage{framed,multirow}

\usepackage[dvipsnames,table]{xcolor} 
\usepackage{xspace}
\usepackage{graphicx}
\usepackage{amssymb}
\usepackage{amsmath}
\usepackage{ dsfont }
\usepackage{multirow}
\definecolor{lightgray}{gray}{0.95}
\definecolor{midgray}{gray}{0.55}
\definecolor{steelblue}{HTML}{4D82B7}
\usepackage{makecell}
\usepackage{nicefrac}
\usepackage[export]{adjustbox}

\newcommand{\dpp}{DER\texttt{++}\xspace}

\newcommand{\resultAF}[2]{\ensuremath{#1}~\scriptsize{(\ensuremath{#2}})}

\newcommand{\resultAFG}[5]{\ensuremath{#1}\textcolor{steelblue}{\textbf{\scriptsize{{$+$#5}}}}~\footnotesize{(\ensuremath{#4})}}

\newcommand{\gainped}[1]{\textcolor{steelblue}{\textbf{\scriptsize{{$+$#1}}}}}

\newcommand{\methodname}{Continual Spectral Regularizer for Incremental Learning\xspace}
\newcommand{\methnam}{CaSpeR-IL\xspace}

\newcommand{\miniimagenet}{\textit{mini}ImageNet\xspace}

\usepackage{amsmath,pict2e}
\DeclareMathOperator*{\argmin}{argmin}

\makeatletter
\newcommand{\barredsum}{%
  \DOTSB\mathop{\mathpalette\@barredsum\relax}\slimits@
}
\newcommand{\@barredsum}[2]{%
  \begingroup
  \sbox\z@{$#1\sum$}%
  \setlength{\unitlength}{\dimexpr2pt+\ht\z@+\dp\z@\relax}%
  \@barredsumthickness{#1}%
  \vphantom{\@barredsumbar}%
  \ooalign{$\m@th#1\sum$\cr\hidewidth$#1\@barredsumbar$\hidewidth\cr}%
  \endgroup
}
\newcommand{\@barredsumbar}{%
  \vcenter{\hbox{\begin{picture}(0,1)\roundcap\Line(0,0)(0,1)\end{picture}}}%
}
\newcommand{\@barredsumthickness}[1]{
  \linethickness{%
    1.25\fontdimen8
      \ifx#1\displaystyle\textfont\else
      \ifx#1\textstyle\textfont\else
      \ifx#1\scriptstyle\scriptfont\else
      \scriptscriptfont\fi\fi\fi 3
  }%
}

\makeatother
\usepackage{hyperref}
\usepackage{enumitem}
\usepackage{algorithm}
\usepackage{algorithmic}

\usepackage{amssymb}
\usepackage{latexsym}

\usepackage{url}
\usepackage{xcolor}
\definecolor{newcolor}{rgb}{.8,.349,.1}

\journal{Pattern Recognition Letters}

\begin{document}


\begin{frontmatter}

\title{Latent Spectral Regularization for Continual Learning}

\author[a1,a2]{Emanuele~\surname{Frascaroli}\corref{cor1}} 
\cortext[cor1]{Corresponding author}
\ead{emanuele.frascaroli@unimore.it}

\author[a1,a2]{Riccardo~\surname{Benaglia}}
\ead{riccardo.benaglia@unimore.it}

\author[a1]{Matteo~\surname{Boschini}}
\ead{matteoboschini3@gmail.com}

\author[a3]{Luca~\surname{Moschella}}
\ead{moschella@di.uniroma1.it}

\author[a1,a2]{Cosimo~\surname{Fiorini}}
\ead{c.fiorini@ammagamma.com}

\author[a3]{Emanuele~\surname{Rodolà}}
\ead{rodola@di.uniroma1.it}

\author[a1]{Simone~\surname{Calderara}}
\ead{simone.calderara@unimore.it}


\affiliation[a1]{organization={University of Modena and Reggio Emilia},
                addressline={Via Vivarelli 10}, 
                city={Modena}, 
                postcode={41125}, 
                country={Italy}}

\affiliation[a2]{organization={Ammagamma},
                addressline={Via S.Orsola 37}, 
                city={Modena}, 
                postcode={41121}, 
                country={Italy}}

\affiliation[a3]{organization={Sapienza University of Rome},
                addressline={Piazzale Aldo Moro 5}, 
                city={Rome}, 
                postcode={00185}, 
                country={Italy}}


\begin{abstract}
While biological intelligence grows organically as new knowledge is gathered throughout life, Artificial Neural Networks forget catastrophically whenever they face a changing training data distribution. Rehearsal-based Continual Learning (CL) approaches have been established as a versatile and reliable solution to overcome this limitation; however, sudden input disruptions and memory constraints are known to alter the consistency of their predictions.
We study this phenomenon by investigating the geometric characteristics of the learner's latent space and find that replayed data points of different classes increasingly mix up, interfering with classification.
Hence, we propose a geometric regularizer that enforces weak requirements on the Laplacian spectrum of the latent space, promoting a partitioning behavior.
Our proposal, called \methodname (\methnam), can be easily combined with any rehearsal-based CL approach and improves the performance of SOTA methods on standard benchmarks.
\end{abstract}


\end{frontmatter}


\section{Introduction}
\label{sec:intro}
Intelligent creatures in the natural world continually learn to adapt their behavior to changing external conditions by seamlessly blending novel notions with previous understanding into a cohesive body of knowledge. In contrast, artificial neural networks (ANNs) greedily fit the data they are currently trained on, swiftly deteriorating previously acquired information, a phenomenon known as \textit{catastrophic forgetting}~\citep{mccloskey1989catastrophic}.
\begin{figure*}[t]
    \centering
    \begin{tabular}{cc}
        \includegraphics[width=.46\linewidth,trim={0 5.05cm 0 0cm},clip]{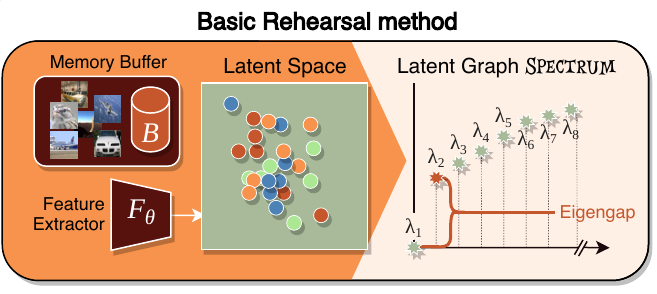} & 
        \includegraphics[width=.46\linewidth,trim={0 0cm 0 5.05cm},clip]{fig_casper.pdf} \\
    \end{tabular}
    
    \caption{An overview of the proposed \methnam regularizer. Rehearsal-based CL methods struggle to separate the latent-space projections of replay data points. Our proposal acts on the spectrum of the latent geometry graph to induce a partitioning behavior by maximizing the \textit{eigengap} for the number of seen classes (\textit{best seen in color}).}
    \label{fig:model}
\end{figure*}
Continual Learning (CL) is a branch of machine learning that designs approaches to help deep models retain previous knowledge while training on new data~\citep{de2019continual}. These methods are evaluated by dividing a classification dataset into disjoint subsets of classes, called \textit{tasks}, letting the model fit one task at a time and evaluating it on all previously seen data~\citep{van2019three}. Recent literature favors the employment of \textit{rehearsal methods}; namely, CL approaches that retain a small memory buffer of samples encountered in previous tasks and interleave them with current training data~\citep{chaudhry2019tiny, buzzega2020dark}.
While rehearsal easily allows the learner to keep track of the joint distribution of all classes seen so far, the limited buffer size produces various overfitting issues that constitute the focus of many recent works (e.g., divergent gradients for new classes~\citep{caccia2022new, boschini2022continual}, deteriorating decision surface~\citep{bonicelli2022effectiveness}, accumulation of predictive bias for current classes~\citep{wu2019large, ahn2020ssilss}).

This paper instead focuses on the changes occurring in the model's latent space as tasks progress. We observe that the learner struggles to separate latent projections of replay examples belonging to different classes, making the downstream classifier prone to interference whenever the input distribution changes and representations are perturbed. Given the Riemannian nature of the latent space of DNNs~\citep{arvanitidis2017latent}, we naturally revert to spectral geometry to study its evolution. Consequently, we introduce a loss term to endow the model's latent space with a cohesive structure without constraining the individual coordinates. As illustrated in Fig.~\ref{fig:model}, our proposed approach, called \textbf{C}ontinu\textbf{a}l \textbf{Spe}ctral \textbf{R}egularizer for \textbf{I}ncremental \textbf{L}earning (\textbf{\methnam}), leverages graph-spectral theory to promote well-separated latent embeddings and can be seamlessly combined with any rehearsal-based CL method to improve its accuracy and robustness against forgetting. 

In summary, we make the following contributions: \textit{i)} we study interference in rehearsal CL models by investigating the geometry of their latent space; \textit{ii)} we propose \methnam: a simple geometrically motivated loss term, inducing the continual learner to produce well-organized latent embeddings; \textit{iii)} we validate our proposal by combining it with several SOTA rehearsal-based CL approaches, showing that \methnam is effective across a wide range of evaluations; \textit{iv)} we compare our work against recent contrastive-based incremental strategies, showing that \methnam better synergizes with CL models; \textit{v)} finally, we present additional studies further investigating the geometric properties conferred by our method on the model's latent space.
The code to reproduce our experiments is available at \href{https://github.com/aimagelab/CaSpeR}{https://github.com/aimagelab/CaSpeR}.

\section{Related Work}
\label{sec:related}
\subsection{Continual Learning}
Continual Learning~\citep{de2019continual} approaches help deep learning models minimize \textit{catastrophic forgetting} when learning on changing input distributions. There are different classes of solutions: \textit{architectural methods} allocate separate portions of the model to separate tasks~\citep{mallya2018packnet}, \textit{regularization methods} use a loss term to prevent changes in the model’s structure or response~\citep{kirkpatrick2017overcoming} and \textit{rehearsal methods} use a working memory buffer to store and replay data-points~\citep{chaudhry2019tiny}
The latter class of approaches is currently the focus of research efforts due to their versatility and effectiveness~\citep{aljundi2019gradient}. Recent trends aim to improve the basic Experience Replay (ER) formula through better memory sampling strategies~\citep{aljundi2019gradient}, combining replay with other optimization techniques~\citep{lopez2017gradient} or providing richer replay signals~\citep{buzzega2020dark}.

A prominent challenge for enhancing \textit{rehearsal methods} is the imbalance between stream and replay data. This can cause a continually learned classifier to struggle to produce unified predictions and be biased towards recently learned classes~\citep{wu2019large}. Researchers have proposed solutions such as architectural modifications of the model~\citep{douillard2020podnet}, alterations to the learning objective of the final classifier~\citep{caccia2022new}, or the use of representation learning instead of cross-entropy~\citep{cha2021co2l}.
Our proposal similarly reduces the intrinsic bias of \textit{rehearsal methods}; it does so by enforcing a desirable property on the model’s latent space through a geometrically motivated regularization term that can be combined with any existing replay method.

A strain of recent CL approaches similarly conditions the model's representation to facilitate clustering by means of a contrastive regularization objective. SCR~\citep{mai2021supervised} enforces consistency between two views of the input batch by leveraging the Supervised Contrastive loss~\citep{khosla2020supervised}; PRD~\citep{asadi2023prd} employs the same loss, in aid to a prototype-based classifier; differently, CSCCT~\citep{ashok2022class} pairs an explicit latent-space clustering objective with a controlled transfer objective preventing negative transfer from dissimilar classes. 
In our experimental section, we compare our proposal against representatives of this family of methods. This allows us to make some interesting observations on how distinct formulations of a similar clustering objective lead to the emergence of different characteristics in latent space geometry.

The use of pre-trained models~\citep{boschini2022transfer, wang2022learning}, also exploiting transformers architectures~\citep{vaswani2017attention}, has been showing increasing popularity in the recent CL literature. We leave the testing of \methnam on those settings for future work, and evaluate our proposal in the more common "train from scratch" scenario.

\begin{figure*}[t]
    \centering
    \begin{tabular}{cc}
        \includegraphics[width=.28\linewidth]{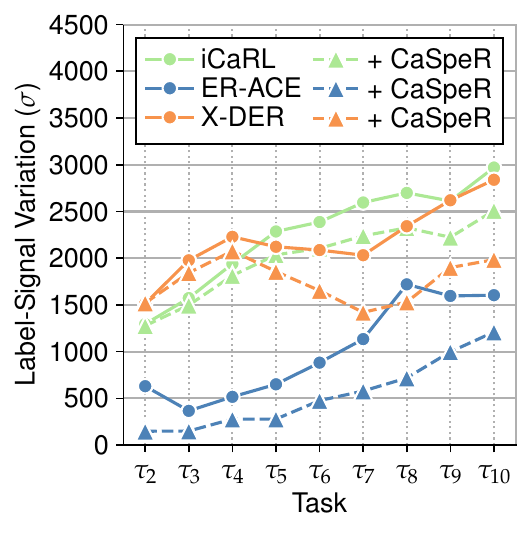} & \includegraphics[width=.65\linewidth]{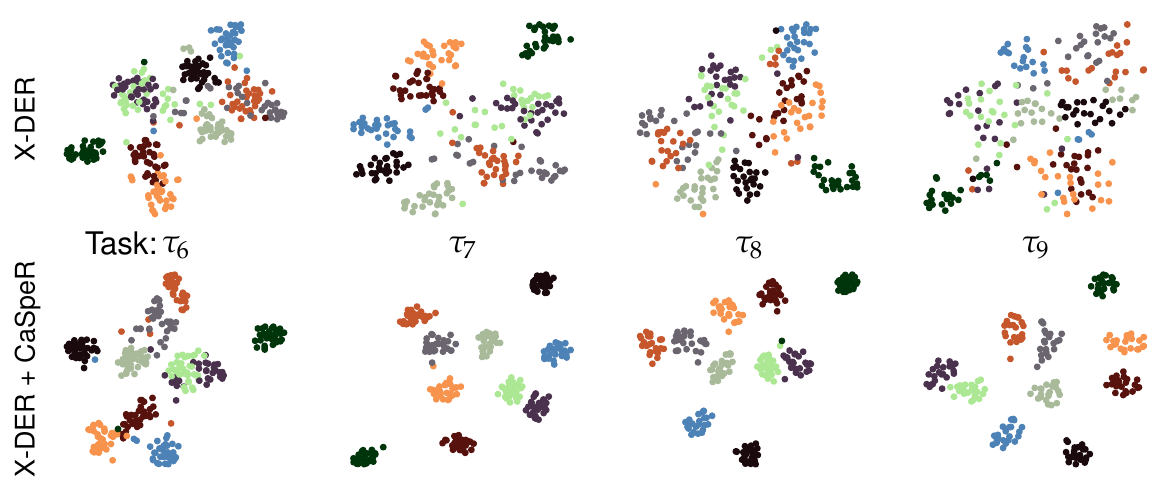} \\
        (a) & (b)
    \end{tabular}
    \caption{How CL alters a model's latent space. (a) A quantitative evaluation measured as Label-Signal Variation ($\sigma$) within the LGG for buffer data points -- \textit{lower is better};
    (b) TSNE embedding of the features computed by X-DER for buffered examples in later tasks (top). Interference between classes is visibly reduced if \methnam is applied (bottom). All experiments are carried out on Split CIFAR-100, (a) uses buffer size $500$, (b) uses $2000$ (\textit{best seen in colors}).}
    \label{fig:ls_qualitative}
\end{figure*}

\subsection{Spectral geometry}
Our approach is built upon the eigendecomposition of the Laplace operator on a graph, falling within the broader area of spectral graph theory. In particular, ours can be regarded as an \emph{inverse} spectral technique, as we prescribe the general behavior of some eigenvalues and seek a graph whose Laplacian spectrum matches this behavior.
In the geometry processing area, such approaches take the name of \emph{isospectralization} techniques and have been recently used in diverse applications such as deformable shape matching~\citep{cosmo2019isospectralization}, shape exploration and reconstruction~\citep{marin2020instant}, shape modeling~\citep{moschella2022learning} and adversarial attacks on shapes~\citep{rampini2021universal}. Differently from these approaches, we work on a single graph (as opposed to pairs of 3D meshes) and our formulation does not take an input spectrum as a target to be matched precisely. Instead, we require the gap between nearby eigenvalues to be maximized, regardless of its exact value. Since our graph represents a discretization of the latent space of a CL model, this simple regularization has important consequences on its learning process.
\section{Method}
\subsection{Continual Learning Setting}
\label{sec:method_continual}
In CL, a learning model $F_{\boldsymbol{\theta}}$ is incrementally exposed to a stream of tasks $\tau_i$, with $i \in \{1, 2, ..., T\}$. The parameters $\boldsymbol{\theta}$ include both the weights of the feature extractor and the classifier, $\boldsymbol{\theta}^f$ and $\boldsymbol{\theta}^c$ respectively.
Each task consists of a sequence of images and their corresponding labels $\tau_i = \{(x^i_1, y^i_1), (x^i_2, y^i_2), ..., (x^i_n, y^i_n)\}$ and does not contain data belonging to classes already seen in previous tasks, so $Y^i \cap Y^j = \text{\O}$, with $i \ne j$ and $Y^i =\{y_k^i\}_{k=1}^{n}$.
At each step $i$, the model cannot freely access data from previous tasks and is optimized by minimizing a loss function $\ell_\text{stream}$ over the current set of examples:
\begin{equation}
\boldsymbol{\theta}^{(i)} = \argmin_{\boldsymbol{\theta}} \ell_\text{stream} =  \argmin_{\boldsymbol{\theta}} \sum_{j=1}^n \ell\big(F_{\boldsymbol{\theta}}(x^i_j), y^i_j\big)\,,
\end{equation}
where the parameters are initialized with the ones obtained after training on the previous task $\boldsymbol{\theta}^{(i-1)}$. If no mechanism is put in place to prevent forgetting, the accuracy on previous tasks will collapse while learning task $\tau_i$~\citep{mccloskey1989catastrophic}. Rehearsal-based CL methods store a pool of examples from previous tasks in a buffer $B$ with fixed size $m$. This data is then used by the model to compute an additional loss term $\ell_{\text{b}}$ aimed at contrasting catastrophic forgetting:
\begin{equation}
\boldsymbol{\theta}^{(i)} = \argmin_{\boldsymbol{\theta}} \ell_\text{stream} + \ell_{\text{b}}.
\end{equation}
For instance, Experience Replay (ER) simply employs a cross-entropy loss over a batch of examples from $B$:
\begin{equation}
\ell_{\text{er}} \triangleq \operatorname{CrossEntropy}\big(F_{\boldsymbol{\theta}}(\boldsymbol{x}^b), \boldsymbol{y}^b\big)\,.
\end{equation}
There exist different strategies for sampling the task data points to fill the buffer. These will be explained in the supplemental material, along with details on the $\ell_{\text{b}}$ employed by each baseline of our experiments.

\subsection{Analysis of changing Latent Space Geometry}
\label{sec:method_geometry}
We are particularly interested in how the latent space changes when introducing a novel task on the input stream. For this reason, we compute the graph $\mathcal{G}$ over the latent-space projection of the replay examples gathered by the CL model after training on $\tau_i$ ($i \in \{2,...T\}$)\footnote{See Sec.~\ref{sec:method_egap} for a detailed description of this procedure.}. 
In order to measure the sparsity of the latent space w.r.t.\ classes representations, we compute the Label-Signal Variation $\sigma$~\citep{lassance2021representing} on the adjacency matrix $\boldsymbol{A} \in \mathbb{R}^{m\times m}$ of $\mathcal{G}$:
\begin{equation}
    \sigma \triangleq \sum_{i=1}^{m} \sum_{j=1}^{m} \mathds{1}_{y^b_i\neq y^b_j} a_{i,j}\,,
\end{equation}
where $\mathds{1}_{\cdot}$ is the indicator function. In Fig.~\ref{fig:ls_qualitative}a, we evaluate several rehearsal CL methods and show they exhibit a steadily growing $\sigma$: examples from distinct classes become more entangled in later tasks. This effect can also be observed qualitatively by considering a TSNE embedding of the points in $B$ (shown in Fig.~\ref{fig:ls_qualitative}b), in which the distances between different-class examples decrease in later tasks. Both evaluations improve when applying our regularizer to the evaluated methods.

\subsection{\methnam: \methodname}
\label{sec:method_egap}
\noindent{\textbf{Motivation.}}~Our method builds upon the fact that the latent spaces of neural models bear a structure informative of the data space they are trained on~\citep{shao2019riemannian}. This structure can be enforced through loss regularizers; \textit{e.g.,} in~\citep{cosmo2020limp}, a minimum-distortion criterion is applied on the latent space of a VAE for a shape generation task. We follow a similar line of thought and propose adopting a geometric term to regularize the latent representations of a CL model.
Namely, we root our approach in spectral geometry; our choice is motivated by the pursuit of a \textbf{compact representation} characterized by \textbf{isometry invariance}. As shown in~\citep{arvanitidis2017latent}, the latent space of DNNs can be modeled as a Riemannian manifold whose \textit{extrinsic} embedding is encoded in the latent vectors. Being extrinsic, these vectors are simply absolute coordinates encoding only one possible realization of the data manifold, out of its infinitely many possible isometries. Each isometry (e.g.; a rotation by 45$^\circ$) would always encode the \textbf{same latent space}, but the \textbf{latent vectors will change} -- this is not desirable, because it may lead to overfitting and lack of generalization. By resorting to spectral geometry, we instead rely on \textit{intrinsic} quantities, that fully encode the latent space and are isometry-invariant.

Our regularizer is based on the graph-theoretic formulation of clustering, where we seek to partition the vertices of $\mathcal{G}$ into well-separated subgraphs with high internal connectivity. A body of results from spectral graph theory, dating back at least to \citep{cheeger1969lower,sinclair1989approximate,shi2000normalized}, explain the gap occurring between neighboring Laplacian eigenvalues as a quantitative measure of graph partitioning. Our proposal, called \methodname (\methnam), draws on these results, but turns the \emph{forward} problem of computing the optimal partitioning of a given graph, into the \emph{inverse} problem of seeking a graph with the desired partitioning.

\noindent{\textbf{Building the LGG.}}~We take the examples in $B$ and forward them through the network; their features are used to build a k-NN graph\footnote{More details on how the k-NN operation can be found in the supplemental material, along with the pseudo-code of \methnam.} $\mathcal{G}$; following \citep{lassance2021representing}, we refer to it as the \textit{latent geometry graph} (LGG).

\noindent{\textbf{Spectral Regularizer.}}~Let us denote by $\boldsymbol{A}$ the adjacency matrix of $\mathcal{G}$, we calculate its degree matrix $\boldsymbol{D}$ and we compute its normalized Laplacian as $
\boldsymbol{L} = \boldsymbol{I} - \boldsymbol{D}^{-\nicefrac{1}{2}} \boldsymbol{A} \boldsymbol{D}^{-\nicefrac{1}{2}}\,$, 
where $\boldsymbol{I}$ is the identity matrix. We then compute the eigenvalues $\boldsymbol{\lambda}$ of $\boldsymbol{L}$ and sort them in ascending order. Let $g$ be the number of different classes within the buffer, we calculate our regularizing loss as:
\begin{equation}
\label{eqn:casperloss}
\ell_{\text{CaSpeR}} \triangleq -\lambda_{g+1} + \sum_{j=1}^g \lambda_j\,.
\end{equation}
The proposed loss term is weighted through the hyperparameter $\rho$ and added to the stream classification loss. Overall, our model optimizes the following objective:
\begin{equation}
    \argmin_{\boldsymbol{\theta}} \ell_{\text{stream}} + \ell_{\text{b}} + \rho \, \ell_{\text{CaSpeR}}\,.
\end{equation}
Through Eq.~\ref{eqn:casperloss}, we increase the eigengap $\lambda_{g+1} - \lambda_g$ while minimizing the first $g$ eigenvalues. Since the number of eigenvalues close to zero represents to the number of loosely connected partitions within the graph~\citep{trevisan14}, our loss indirectly encourages data points to be clustered without strict supervision.

\noindent{\textbf{Efficient Batch Operation.}}~The application of \methnam entails the cumbersome step of constructing the entire LGG $\mathcal{G}$ at each forward step by processing all available replay examples in the buffer $B$ (usually orders of magnitude larger than a batch of input examples).
We consequently propose an efficient approximation of our initial objective by not operating on $\mathcal{G}$ directly, but rather sampling a random sub-graph $\mathcal{G}_\text{p} \subset \mathcal{G}$ spanning only $p$ out of the $g$ classes represented in the memory buffer. As $\mathcal{G}_\text{p}$ still includes a conspicuous amount of nodes, we resort to an additional sub-sampling and extract $\mathcal{G}_\text{p}^t \subset \mathcal{G}_\text{p}$, a smaller graph with $t$ exemplars for each class.
By repeating these random samplings in each forward step, we optimize a Monte Carlo approximation of Eq.~\ref{eqn:casperloss}:
\begin{equation}
    \ell_{\text{CaSpeR}}^* \triangleq \mathop{\mathbb{E}}_{\mathcal{G}_\text{p} \subset \mathcal{G}}\Bigg[ \mathop{\mathbb{E}}_{\mathcal{G}_\text{p}^t \subset \mathcal{G}_\text{p}} \bigg[ -\lambda^{\mathcal{G}_\text{p}^t}_{p+1} + \sum_{j=1}^p \lambda^{\mathcal{G}_\text{p}^t}_j\bigg]\Bigg]\,,
\end{equation}
where the $\boldsymbol{\lambda}^{\mathcal{G}_\text{p}^t}$ denote the eigenvalues of the Laplacian of $\mathcal{G}_\text{p}^t$. Here, we enforce the eigengap at $p$, as we know by construction that each $\mathcal{G}_\text{p}^t$ comprises samples from $p$ communities within $\mathcal{G}$.
In practice, we extract $b$ samples from the buffer, maintaining $b$ equal to the batch size to ensure balance~\citep{buzzega2020dark,buzzega2020rethinking}. This results in $t=\frac{b}{p}$ samples from each class of the $p$ randomly chosen. An in-depth discussion on the hyperparameters of \methnam can be found in the Supplemental Material.
\section{Continual Learning Experiments}
\label{sec:exps}
\subsection{Evaluation}
\label{sec:exp_proto}
\noindent\textbf{Settings.}~To assess the effectiveness of the proposed method,
we prioritize \textit{Class Incremental Learning} (Class-IL) classification protocol~\citep{van2019three}, where the model learns to make predictions in the absence of task information, as it is recognized as a more realistic and challenging benchmark~\citep{farquhar2018towards, aljundi2019gradient}. In the supplemental material, we report results for both \textit{Task-Incremental Learning} (Task-IL) and \textit{Domain-Incremental Learning} (Domain-IL) protocols, demonstrating that \methnam can enhance CL baselines within these scenarios as well.

\noindent\textbf{Benchmarked models.}~To evaluate the benefit of our regularizer, we apply it on top of several SOTA rehearsal-based methods: Experience Replay with Asymmetric Cross-Entropy \textbf{(ER-ACE)}~\citep{caccia2022new}, Incremental Classifier and Representation Learning \textbf{(iCaRL)}~\citep{rebuffi2017icarl}, Dark Experience Replay \textbf{(DER++)}~\citep{buzzega2020dark}, eXtended-DER \textbf{(X-DER)}
and Pooled Outputs Distillation Network \textbf{(PODNet)}~\citep{douillard2020podnet}.

\begin{table*}[t]
\caption{Class-IL results -- $\bar{A}_F$ ($\bar{F}^*_F$) -- for SOTA rehearsal CL methods, with and without \methnam.} \label{tab:cil}

\begin{center}
\setlength{\tabcolsep}{2pt}
\rowcolors{5}{lightgray}{}
\begin{tabular}{l@{\hskip 0.3cm}cc@{\hskip 0.3cm}cc@{\hskip 0.3cm}cc}
\hline
\textbf{Class-IL} & \multicolumn{2}{c}{\textbf{Split CIFAR-10}} & \multicolumn{2}{c}{\textbf{Split CIFAR-100}} & \multicolumn{2}{c}{\textbf{Split \miniimagenet}}\\
\hline
Joint (UB)        & \multicolumn{2}{c}{\resultAF{87.08}{-}}                  & \multicolumn{2}{c}{\resultAF{63.11}{-}}               & \multicolumn{2}{c}{\resultAF{52.76}{-}} \\
Finetune (LB)        & \multicolumn{2}{c}{\resultAF{19.53}{100.00}}               & \multicolumn{2}{c}{\resultAF{8.38}{100.00}}            & \multicolumn{2}{c}{\resultAF{3.87}{100.00}} \\
\hline
\textbf{Buffer Size} & 500 & 1000 & 500 & 2000 & 2000 & 5000\\
\hline

ER-ACE  & \resultAF{66.13}{21.76} & \resultAF{71.72}{14.88} & \resultAF{34.99}{51.41} & \resultAF{46.52}{34.60} & \resultAF{22.03}{49.04} & \resultAF{27.26}{29.99} \\
\hspace{.3em} + \textbf{\methnam}  & \resultAF{69.58}{20.56} & \resultAF{73.82}{14.11} & \resultAF{36.70}{46.61} & \resultAF{47.85}{33.86} & \resultAF{23.36}{47.90} & \resultAF{29.15}{28.36} \\

iCaRL & \resultAF{52.71}{22.69} & \resultAF{62.94}{21.64} & \resultAF{39.56}{32.73} & \resultAF{40.47}{31.24} & \resultAF{19.42}{36.89} & \resultAF{20.17}{33.23} \\
\hspace{.3em} + \textbf{\methnam} & \resultAF{55.66}{20.56} & \resultAF{63.99}{21.05} & \resultAF{40.87}{32.31} & \resultAF{41.83}{25.55} & \resultAF{20.46}{35.90} & \resultAF{21.45}{32.26} \\

\dpp & \resultAF{67.38}{26.77} & \resultAF{71.17}{25.12} & \resultAF{28.01}{57.56} & \resultAF{43.27}{34.94} & \resultAF{20.88}{74.48} & \resultAF{28.55}{61.03} \\
\hspace{.3em} + \textbf{\methnam} & \resultAF{69.11}{26.18} & \resultAF{73.12}{23.43} & \resultAF{32.16}{53.41} & \resultAF{46.95}{30.08} & \resultAF{22.61}{71.01} & \resultAF{29.96}{57.60} \\

X-DER & \resultAF{63.23}{14.99} & \resultAF{65.72}{12.28} & \resultAF{35.89}{44.54} & \resultAF{46.37}{23.57} & \resultAF{24.80}{44.69} & \resultAF{30.98}{30.12} \\
\hspace{.3em} + \textbf{\methnam} & \resultAF{65.56}{14.41} & \resultAF{67.84}{10.65} & \resultAF{38.23}{43.90} & \resultAF{48.11}{18.47} & \resultAF{26.24}{41.72} & \resultAF{31.63}{28.71} \\

PODNet & \resultAF{37.22}{40.49} & \resultAF{45.97}{39.49} & \resultAF{30.16}{54.49} & \resultAF{32.12}{46.73} & \resultAF{16.82}{52.32} & \resultAF{20.81}{46.50} \\
\hspace{.3em} + \textbf{\methnam} & \resultAF{39.85}{39.51} & \resultAF{47.40}{38.90} & \resultAF{32.27}{48.32} & \resultAF{38.64}{35.65} & \resultAF{18.09}{50.33} & \resultAF{23.63}{45.08} \\
\hline
\end{tabular}
\end{center}
\end{table*}
We include the performance of the upper bound training on all classes together in an offline manner (Joint) and the lower bound training on each task sequentially without any method to prevent forgetting (Finetune).

\noindent\textbf{Datasets.}~We conduct the experiments on three commonly used image datasets, splitting the classes from the main dataset into separate disjoint sets used to sequentially train the evaluated models.
For \textbf{Split CIFAR-10} we adopt the standard benchmark of splitting the dataset into $5$ subsets of $2$ classes each;
for \textbf{Split CIFAR-100}, we exploit the 100-class CIFAR100~\citep{krizhevsky2009learning} dataset by splitting the dataset into $10$ subsets of $10$ classes each;
for \textbf{Split \miniimagenet}, we leverage the \miniimagenet~\citep{vinyals2016matching} Imagenet subset, adopting the $20$ tasks per $5$ classes protocol.

\noindent\textbf{Metrics.}~We mainly quantify the performance of the compared models in terms of \textit{Final Average Accuracy} $
\bar{A}_F \triangleq
\frac{1}{T}\sum_{i=1}^{T}{a_i^T}$,
where $a_i^j$ is the accuracy of the model at the end of task $j$ calculated on the test set of task $\tau_i$ and reported in percentage value.
To quantify the severity of the performance degradation that occurs as a result of catastrophic forgetting, we exploit \textit{Final Average Adjusted Forgetting} ($\bar{F}^*_F$), as defined in~\citep{bonicelli2023raider}. It is a $[0, 100]$-bounded version of the popular forgetting metric~\citep{chaudhry2018riemannian}. 

\noindent\textbf{Hyperparameter selection.}~To ensure a fair evaluation, we train all the models with the same batch size and the same number of epochs. Moreover, we employ the same backbone for all experiments on the same dataset. In particular, we use Resnet18 \citep{he2016deep} for Split CIFAR-100 and Split CIFAR-10 and EfficientNet-B2 \citep{tan2019efficientnet} for Split \miniimagenet. The best hyperparameters for each model-dataset configuration are found via grid search.
For additional details and further experiments with varying training epochs and batch sizes, demonstrating the effectiveness of \methnam under different conditions, we direct the reader to the supplemental material.

\subsection{Results}
\label{sec:results_comment}
We report a breakdown of Class-IL results of our evaluation in Tab.~\ref{tab:cil}.  
\methnam leads to a firm improvement in $\bar{A}_F$ across all evaluated methods and datasets. The steady reduction of $\bar{F}^*_F$ confirms that the regularization adopted effectively addresses catastrophic forgetting.

We notice that the improvement in accuracy does not always grow with the memory buffer size. This is in contrast with the typical behavior of replay regularization terms~\citep{cha2021co2l,chaudhry2019tiny}. We believe it is due to our distinctively geometric approach: as spectral properties of graphs are understood to be robust w.r.t.~coarsening~\citep{coarsening}, \methnam does not need a large pool of data to be effective.

In Task-IL and Domain-IL (results in the supplemental material), the gains are lower than in Class-IL: existing methods are already strong in these less challenging scenarios. However, \methnam still provides a steady improvement, proving its ability to both consolidate the knowledge of each task individually (Task-IL) and counteract the bias introduced by new data distribution on known classes (Domain-IL).

\begin{figure*}[t]
    \centering
    \includegraphics[width=0.95\linewidth]{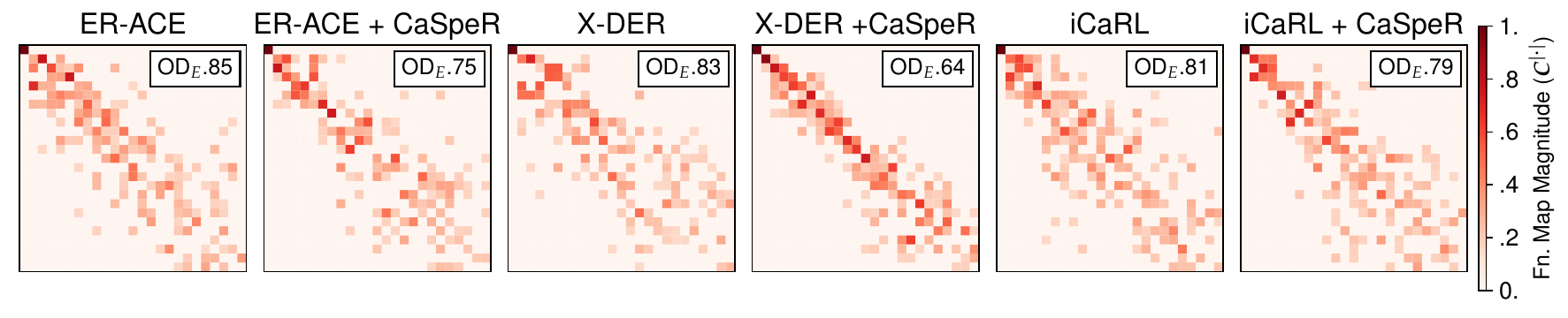}
    \caption{For several rehearsal methods with and without \methnam, the functional map magnitude matrices $\boldsymbol{C^{|\cdot|}}$ between the LGGs $\mathcal{G}^{\tau_5}$ and $\mathcal{G}^{\tau_{10}}$, computed on the test set of $\tau_1,...,\tau_5$ after training up to $\tau_5$ and $\tau_{10}$ respectively (Split CIFAR-100 - buffer size $2000$). The closer $\boldsymbol{C^{|\cdot|}}$ to the diagonal, the less geometric distortion between $\mathcal{G}^{\tau_5}$ and $\mathcal{G}^{\tau_{10}}$. We report the first $25$ rows and columns of $\boldsymbol{C^{|\cdot|}}$, focusing on low-frequency correspondences~\citep{ovsjanikov2012functional}, and apply a $\boldsymbol{C^{|\cdot|}} > 0.15$ threshold to increase clarity.}
    \label{fig:fmaps}
\end{figure*}

\begin{table}[t]
    \caption{Comparison with contrastive baselines. We report $\bar{A}_F$ and the average variance of same-class projections on the latent space.}
    \label{tab:scr}

\centering
\setlength{\tabcolsep}{2pt}
\rowcolors{5}{lightgray}{}
\begin{tabular}{l@{\hskip 0.3cm}cc@{\hskip 0.3cm}cc}
\hline
\textbf{Class-IL} &  \multicolumn{4}{c}{\textbf{Split CIFAR-100}} \\
\hline
\textbf{Buffer Size} & \multicolumn{2}{c@{\hskip 0.3cm}}{500} & \multicolumn{2}{c}{2000} \\
\hline
 & $\bar{A}_F$ & Variance & $\bar{A}_F$ & Variance \\
\hline
SCR                                 & $31.18$ & $2.2111$ & $43.39$ & $4.4439$ \\
\hline

ER-ACE                              & $34.99$ & $0.5313$ & $46.52$ & $0.5769$ \\
\hspace{.3em} + \textbf{\methnam}   & $\boldsymbol{36.70}$ & $0.4926$ & $\boldsymbol{47.85}$ & $0.5478$ \\
\hspace{.3em} + \textbf{CSCCT}      & $34.93$ & $\boldsymbol{0.3931}$ & $45.91$ & $\boldsymbol{0.4290}$ \\

\hline

iCaRL                               & $39.56$ & $0.8381$ & $40.47$ & $0.8248$ \\
\hspace{.3em} + \textbf{\methnam}   & $\boldsymbol{40.57}$ & $\boldsymbol{0.8289}$ & $\boldsymbol{41.83}$ & $\boldsymbol{0.8057}$ \\
\hspace{.3em} + \textbf{CSCCT}      & $39.36$ & $0.9167$ & $40.87$ & $1.0392$ \\

\hline

\dpp                                & $28.01$ & $0.1283$ & $43.27$ & $0.1209$ \\
\hspace{.3em} + \textbf{\methnam}   & $\boldsymbol{32.16}$ & $0.0964$ & $\boldsymbol{46.95}$ & $0.1012$ \\
\hspace{.3em} + \textbf{CSCCT}      & $30.17$ & $\boldsymbol{0.0552}$ & $44.27$ & $\boldsymbol{0.0857}$ \\

\hline

X-DER                               & $35.89$ & $0.2265$ & $46.37$ & $0.2523$ \\
\hspace{.3em} + \textbf{\methnam}   & $\boldsymbol{38.23}$ & $0.2065$ & $\boldsymbol{48.11}$ & $\boldsymbol{0.2207}$ \\
\hspace{.3em} + \textbf{CSCCT}      & $36.23$ & $\boldsymbol{0.1974}$ & $45.51$ & $0.2242$ \\

\hline

PODNet                              & $30.16$ & $0.4229$ & $32.12$ & $0.7366$ \\
\hspace{.3em} + \textbf{\methnam}   & $\boldsymbol{32.27}$ & $0.4197$ & $\boldsymbol{38.64}$ & $0.5700$ \\
\hspace{.3em} + \textbf{CSCCT}      & $30.78$ & $\boldsymbol{0.1809}$ & $33.59$ & $\boldsymbol{0.2577}$ \\

\end{tabular}

\end{table}

\noindent\textbf{Comparison with Contrastive Learning.}
The thorough study in \citep{haochen2021provable} interprets contrastive learning as a parametric form of spectral clustering on the input augmentation graph, which points to a link with our approach. Given the similarity between \methnam's goal and contrastive objectives, we devise a comparison with SCR (Supervised Contrastive Replay)~\citep{khosla2020supervised} and CSCCT (Cross-Space Clustering and Controlled Transfer)~\citep{ashok2022class}, two existing CL contrastive baselines described in Sec.~\ref{sec:related}.
We evaluated these methods on the Split CIFAR-100 benchmark described above. 
SCR is a standalone model extending Experience Replay;  conversely, CSCCT is a module that can be plugged into existing CL methods, so we consider it as a direct competitor implemented upon our baselines. Despite behaving similarly to \methnam, CSCCT requires a past model snapshot to be available during training and inserts both streaming and memory data within its loss terms.
Results can be seen in Tab.~\ref{tab:scr}. To better analyze the effects of different CL approaches on the latent space, we measured the average variance of same-class projections at the end of training.

Firstly, we observe that SCR exhibits a higher variance in the latent space compared to other baselines. 
Conversely, both CSCCT and \methnam are able to reduce the latent-space variance of the model they are applied to. The results highlight an intriguing behavior: despite CSCCT often achieving minimal variance, the accuracy improvement of \methnam remains higher. Hence, we suggest two interesting explanations for these observations: \textit{i)} intra-class variance in latent space is not proportional to accuracy; \textit{ii)} while directly constraining individual coordinates may limit the model’s ability to rearrange data points, spectral geometry may be a softer clustering approach that permits the flexibility required to organize the latent space into more optimal structures for classification tasks.

\begin{table}[t]
    \caption{Class-IL $\bar{A}_F$ values of k-NN classifiers trained on top of the latent representations of replay data points. Results on Split CIFAR-100 for Buffer Size 2000.}
    \label{tab:knn} 
    \begin{center}
\rowcolors{3}{lightgray}{}
\begin{tabular}{lcccc}
\hline
\textbf{k-NN Clsf} & \multicolumn{2}{c}{w/o \methnam} & \multicolumn{2}{c}{w/ \methnam}\\
\textbf{(Class-IL)}         & 5-NN & 11-NN & 5-NN & 11-NN\\
\hline
ER-ACE                              & $43.73$ & $44.41$ & $46.75$\gainped{3.02} & $47.29$\gainped{2.88} \\
iCaRL                               & $34.86$ & $37.78$ & $36.00$\gainped{1.14} & $38.33$\gainped{0.55} \\
\dpp                                & $44.21$ & $44.24$ & $45.75$\gainped{1.54} & $46.00$\gainped{1.76} \\
X-DER                               & $43.44$ & $44.62$ & $49.47$\gainped{6.03} & $49.49$\gainped{4.87} \\
PODNet                              & $21.11$ & $22.60$ & $27.88$\gainped{6.77} & $28.94$\gainped{6.34} \\
\hline
\end{tabular}
\end{center}

\end{table}

\section{Model Analysis}

\subsection{k-NN classification}
To further verify whether \methnam successfully separates the latent embeddings for examples of different classes, we evaluate the accuracy of k-NN-classifiers~\citep{wu2018unsupervised} trained on top of the latent representations produced by the methods of Sec.~\ref{sec:exps}. In Tab.~\ref{tab:knn}, we report the results for 5-NN and 11-NN classifiers using the final buffer $B$ as a support set. We observe that \methnam also shows its steady beneficial effect on top of this classification approach, further confirming its validity in disentangling the representations of different classes.

\subsection{Latent Space Consistency}
To provide further insights into the dynamics of the latent space on the evaluated models, we study the emergence of distortions in the LGG. Given a continual learning model, we are interested in a comparison between $\mathcal{G}^{\tau_5}$ and $\mathcal{G}^{\tau_{10}}$, the LGGs produced after training on $\tau_5$ and $\tau_{10}$ respectively, computed on the test set of tasks $\tau_1,...,\tau_5$.

The comparison between $\mathcal{G}^{\tau_5}$ and $\mathcal{G}^{\tau_{10}}$ can be better understood in terms of the node-to-node bijection $T:\mathcal{G}^{\tau_5} \rightarrow \mathcal{G}^{\tau_{10}}$, which can be represented as a functional map matrix $\boldsymbol{C}$~\citep{ovsjanikov2012functional} with elements $c_{i,j} \triangleq \langle \boldsymbol{\phi}_i^{\mathcal{G}^{\tau_5}} , \boldsymbol{\phi}_j^{\mathcal{G}^{\tau_{10}}} \circ T\rangle\,$,
where $\boldsymbol{\phi}_i^{\mathcal{G}^{\tau_5}}$ is the $i$-th Laplacian eigenvector of $\mathcal{G}^{\tau_5}$ (similarly for $\mathcal{G}^{\tau_{10}}$), and $\circ$ denotes the standard function composition. In other words, the matrix $\boldsymbol{C}$ encodes the similarity between the Laplacian eigenspaces of the two graphs.
In an ideal scenario where the latent space is subject to no modification between $\tau_5$ and $\tau_{10}$ w.r.t.\ previously learned classes, $T$ is an \textit{isomorphism} and $\boldsymbol{C}$ is a diagonal matrix~\citep{ovsjanikov2012functional}. In a practical scenario, $T$ is only approximately isomorphic, and, the better the approximation, the more $\boldsymbol{C}$ is sparse and funnel-shaped.

In Fig.~\ref{fig:fmaps}, we report $\boldsymbol{C^{|\cdot|}}\triangleq \operatorname{abs}(\boldsymbol{C})$ for ER-ACE, iCaRL and X-DER on Split CIFAR-100, both with and without \methnam. It can be observed that the methods that benefit from our proposal display a tighter functional map matrix. This indicates that the partitioning behavior promoted by \methnam leads to reduced interference, as the portion of the LGG that refers to previously learned classes remains geometrically consistent in later tasks. 
To quantify the similarity of each $\boldsymbol{C^{|\cdot|}}$ matrix to the identity, we also report its off-diagonal energy $OD_{E}$~\citep{rodola2017partial}, computed as the sum of the elements outside of the main diagonal divided by the Frobenius norm.
\methnam produces a clear decrease in $OD_{E}$, signifying an increase in the diagonality of the functional matrices.
\subsection{Limitations}
Given the necessity for the proposed regularizer to store and reuse previously learned training samples, we remark that \methnam applicability might be limited if privacy constraints are in place. 
This applies to any rehearsal CL method.
\section{Conclusion}
In this work, we investigate the evolution of a CL model's latent space throughout training. We find that latent-space projections of past exemplars are relentlessly drawn closer together, paving the way for catastrophic forgetting.
Drawing on spectral graph theory, we propose \methnam: a regularizer that encourages the clustering of data points in the latent space, without constraining individual coordinates. We show that our approach can be easily combined with any rehearsal-based CL approach, improving their performance on standard benchmarks.

\section*{Acknowledgments}
This paper has been supported from Italian Ministerial grant PRIN 2020 ``LEGO.AI: LEarning the Geometry of knOwledge in AI systems'', n. 2020TA3K9N. We acknowledge the CINECA award under the ISCRA initiative, for the availability of high performance computing resources and support.


\bibliographystyle{model5-names}
\bibliography{bib_compact,bib_bibliography}

\begin{thebibliography}{52}
\expandafter\ifx\csname natexlab\endcsname\relax\def\natexlab#1{#1}\fi
\providecommand{\url}[1]{\texttt{#1}}
\providecommand{\href}[2]{#2}
\providecommand{\path}[1]{#1}
\providecommand{\DOIprefix}{doi:}
\providecommand{\ArXivprefix}{arXiv:}
\providecommand{\URLprefix}{URL: }
\providecommand{\Pubmedprefix}{pmid:}
\providecommand{\doi}[1]{\href{https://doi.org/#1}{\path{#1}}}
\providecommand{\Pubmed}[1]{\href{pmid:#1}{\path{#1}}}
\providecommand{\bibinfo}[2]{#2}
\ifx\xfnm\relax \def\xfnm[#1]{\unskip,\space#1}\fi
\bibitem[{Ahn et~al.(2021)Ahn, Kwak, Lim, Bang, Kim \& Moon}]{ahn2020ssilss}
\bibinfo{author}{Ahn, H.}, \bibinfo{author}{Kwak, J.}, \bibinfo{author}{Lim,
  S.}, \bibinfo{author}{Bang, H.}, \bibinfo{author}{Kim, H.}, \&
  \bibinfo{author}{Moon, T.} (\bibinfo{year}{2021}).
\newblock \bibinfo{title}{{SS-IL: Separated Softmax for Incremental Learning.}}
\newblock In {\it \bibinfo{booktitle}{ICCV}\/}.
\bibitem[{Aljundi et~al.(2019)Aljundi, Lin, Goujaud \&
  Bengio}]{aljundi2019gradient}
\bibinfo{author}{Aljundi, R.}, \bibinfo{author}{Lin, M.},
  \bibinfo{author}{Goujaud, B.}, \& \bibinfo{author}{Bengio, Y.}
  (\bibinfo{year}{2019}).
\newblock \bibinfo{title}{{Gradient Based Sample Selection for Online Continual
  Learning}}.
\newblock In {\it \bibinfo{booktitle}{ANeurIPS}\/}.
\bibitem[{Arvanitidis et~al.(2018)Arvanitidis, Hansen \&
  Hauberg}]{arvanitidis2017latent}
\bibinfo{author}{Arvanitidis, G.}, \bibinfo{author}{Hansen, L.~K.}, \&
  \bibinfo{author}{Hauberg, S.} (\bibinfo{year}{2018}).
\newblock \bibinfo{title}{Latent space oddity: on the curvature of deep
  generative models}.
\newblock In {\it \bibinfo{booktitle}{ICLR}\/}.
\bibitem[{Asadi et~al.(2023)Asadi, Davari, Mudur, Aljundi \&
  Belilovsky}]{asadi2023prd}
\bibinfo{author}{Asadi, N.}, \bibinfo{author}{Davari, M.},
  \bibinfo{author}{Mudur, S.}, \bibinfo{author}{Aljundi, R.}, \&
  \bibinfo{author}{Belilovsky, E.} (\bibinfo{year}{2023}).
\newblock \bibinfo{title}{Prototype-sample relation distillation: towards
  replay-free continual learning}.
\newblock In {\it \bibinfo{booktitle}{ICLR}\/}.
\bibitem[{Ashok et~al.(2022)Ashok, Joseph \& Balasubramanian}]{ashok2022class}
\bibinfo{author}{Ashok, A.}, \bibinfo{author}{Joseph, K.}, \&
  \bibinfo{author}{Balasubramanian, V.~N.} (\bibinfo{year}{2022}).
\newblock \bibinfo{title}{Class-incremental learning with cross-space
  clustering and controlled transfer}.
\newblock In {\it \bibinfo{booktitle}{ECCV}\/}.
\bibitem[{Bonicelli et~al.(2023)Bonicelli, Boschini, Frascaroli, Porrello,
  Pennisi, Bellitto, Palazzo, Spampinato \& Calderara}]{bonicelli2023raider}
\bibinfo{author}{Bonicelli, L.}, \bibinfo{author}{Boschini, M.},
  \bibinfo{author}{Frascaroli, E.}, \bibinfo{author}{Porrello, A.},
  \bibinfo{author}{Pennisi, M.}, \bibinfo{author}{Bellitto, G.},
  \bibinfo{author}{Palazzo, S.}, \bibinfo{author}{Spampinato, C.}, \&
  \bibinfo{author}{Calderara, S.} (\bibinfo{year}{2023}).
\newblock \bibinfo{title}{On the effectiveness of equivariant regularization
  for robust online continual learning}.
\newblock {\it \bibinfo{journal}{arXiv preprint arXiv:2305.03648}\/}, .
\bibitem[{Bonicelli et~al.(2022)Bonicelli, Boschini, Porrello, Spampinato \&
  Calderara}]{bonicelli2022effectiveness}
\bibinfo{author}{Bonicelli, L.}, \bibinfo{author}{Boschini, M.},
  \bibinfo{author}{Porrello, A.}, \bibinfo{author}{Spampinato, C.}, \&
  \bibinfo{author}{Calderara, S.} (\bibinfo{year}{2022}).
\newblock \bibinfo{title}{{On the Effectiveness of Lipschitz-Driven Rehearsal
  in Continual Learning}}.
\newblock In {\it \bibinfo{booktitle}{ANeurIPS}\/}.
\bibitem[{Boschini et~al.(2022{\natexlab{a}})Boschini, Bonicelli, Buzzega,
  Porrello \& Calderara}]{boschini2022class}
\bibinfo{author}{Boschini, M.}, \bibinfo{author}{Bonicelli, L.},
  \bibinfo{author}{Buzzega, P.}, \bibinfo{author}{Porrello, A.}, \&
  \bibinfo{author}{Calderara, S.} (\bibinfo{year}{2022}{\natexlab{a}}).
\newblock \bibinfo{title}{Class-incremental continual learning into the
  extended der-verse}.
\newblock {\it \bibinfo{journal}{IEEE TPAMI}\/}, .
\bibitem[{Boschini et~al.(2022{\natexlab{b}})Boschini, Bonicelli, Porrello,
  Bellitto, Pennisi, Palazzo, Spampinato \& Calderara}]{boschini2022transfer}
\bibinfo{author}{Boschini, M.}, \bibinfo{author}{Bonicelli, L.},
  \bibinfo{author}{Porrello, A.}, \bibinfo{author}{Bellitto, G.},
  \bibinfo{author}{Pennisi, M.}, \bibinfo{author}{Palazzo, S.},
  \bibinfo{author}{Spampinato, C.}, \& \bibinfo{author}{Calderara, S.}
  (\bibinfo{year}{2022}{\natexlab{b}}).
\newblock \bibinfo{title}{Transfer without forgetting}.
\newblock In {\it \bibinfo{booktitle}{ECCV}\/}.
\bibitem[{Boschini et~al.(2022{\natexlab{c}})Boschini, Buzzega, Bonicelli,
  Porrello \& Calderara}]{boschini2022continual}
\bibinfo{author}{Boschini, M.}, \bibinfo{author}{Buzzega, P.},
  \bibinfo{author}{Bonicelli, L.}, \bibinfo{author}{Porrello, A.}, \&
  \bibinfo{author}{Calderara, S.} (\bibinfo{year}{2022}{\natexlab{c}}).
\newblock \bibinfo{title}{Continual semi-supervised learning through
  contrastive interpolation consistency}.
\newblock {\it \bibinfo{journal}{PRL}\/}, .
\bibitem[{Buzzega et~al.(2020{\natexlab{a}})Buzzega, Boschini, Porrello, Abati
  \& Calderara}]{buzzega2020dark}
\bibinfo{author}{Buzzega, P.}, \bibinfo{author}{Boschini, M.},
  \bibinfo{author}{Porrello, A.}, \bibinfo{author}{Abati, D.}, \&
  \bibinfo{author}{Calderara, S.} (\bibinfo{year}{2020}{\natexlab{a}}).
\newblock \bibinfo{title}{{Dark Experience for General Continual Learning: a
  Strong, Simple Baseline}}.
\newblock In {\it \bibinfo{booktitle}{ANeurIPS}\/}.
\bibitem[{Buzzega et~al.(2020{\natexlab{b}})Buzzega, Boschini, Porrello \&
  Calderara}]{buzzega2020rethinking}
\bibinfo{author}{Buzzega, P.}, \bibinfo{author}{Boschini, M.},
  \bibinfo{author}{Porrello, A.}, \& \bibinfo{author}{Calderara, S.}
  (\bibinfo{year}{2020}{\natexlab{b}}).
\newblock \bibinfo{title}{{Rethinking Experience Replay: a Bag of Tricks for
  Continual Learning}}.
\newblock In {\it \bibinfo{booktitle}{ICPR}\/}.
\bibitem[{Caccia et~al.(2022)Caccia, Aljundi, Asadi, Tuytelaars, Pineau \&
  Belilovsky}]{caccia2022new}
\bibinfo{author}{Caccia, L.}, \bibinfo{author}{Aljundi, R.},
  \bibinfo{author}{Asadi, N.}, \bibinfo{author}{Tuytelaars, T.},
  \bibinfo{author}{Pineau, J.}, \& \bibinfo{author}{Belilovsky, E.}
  (\bibinfo{year}{2022}).
\newblock \bibinfo{title}{{New Insights on Reducing Abrupt Representation
  Change in Online Continual Learning}}.
\newblock In {\it \bibinfo{booktitle}{ICLR}\/}.
\bibitem[{Cha et~al.(2021)Cha, Lee \& Shin}]{cha2021co2l}
\bibinfo{author}{Cha, H.}, \bibinfo{author}{Lee, J.}, \& \bibinfo{author}{Shin,
  J.} (\bibinfo{year}{2021}).
\newblock \bibinfo{title}{Co2l: Contrastive continual learning}.
\newblock In {\it \bibinfo{booktitle}{ICCV}\/}.
\bibitem[{Chaudhry et~al.(2018)Chaudhry, Dokania, Ajanthan \&
  Torr}]{chaudhry2018riemannian}
\bibinfo{author}{Chaudhry, A.}, \bibinfo{author}{Dokania, P.~K.},
  \bibinfo{author}{Ajanthan, T.}, \& \bibinfo{author}{Torr, P.~H.}
  (\bibinfo{year}{2018}).
\newblock \bibinfo{title}{{Riemannian walk for incremental learning:
  Understanding forgetting and intransigence}}.
\newblock In {\it \bibinfo{booktitle}{ECCV}\/}.
\bibitem[{Chaudhry et~al.(2019)Chaudhry, Rohrbach, Elhoseiny, Ajanthan,
  Dokania, Torr \& Ranzato}]{chaudhry2019tiny}
\bibinfo{author}{Chaudhry, A.}, \bibinfo{author}{Rohrbach, M.},
  \bibinfo{author}{Elhoseiny, M.}, \bibinfo{author}{Ajanthan, T.},
  \bibinfo{author}{Dokania, P.~K.}, \bibinfo{author}{Torr, P.~H.}, \&
  \bibinfo{author}{Ranzato, M.} (\bibinfo{year}{2019}).
\newblock \bibinfo{title}{{On tiny episodic memories in continual learning}}.
\newblock In {\it \bibinfo{booktitle}{ICML Workshops}\/}.
\bibitem[{Cheeger(1969)}]{cheeger1969lower}
\bibinfo{author}{Cheeger, J.} (\bibinfo{year}{1969}).
\newblock \bibinfo{title}{A lower bound for the smallest eigenvalue of the
  laplacian}.
\newblock In {\it \bibinfo{booktitle}{Problems in analysis}\/}.
\newblock \bibinfo{publisher}{Princeton University Press}.
\bibitem[{Cosmo et~al.(2020)Cosmo, Norelli, Halimi, Kimmel \&
  Rodol\`{a}}]{cosmo2020limp}
\bibinfo{author}{Cosmo, L.}, \bibinfo{author}{Norelli, A.},
  \bibinfo{author}{Halimi, O.}, \bibinfo{author}{Kimmel, R.}, \&
  \bibinfo{author}{Rodol\`{a}, E.} (\bibinfo{year}{2020}).
\newblock \bibinfo{title}{Limp: Learning latent shape representations with
  metric preservation priors}.
\newblock In {\it \bibinfo{booktitle}{ECCV}\/}.
\bibitem[{Cosmo et~al.(2019)Cosmo, Panine, Rampini, Ovsjanikov, Bronstein \&
  Rodol{\`{a}}}]{cosmo2019isospectralization}
\bibinfo{author}{Cosmo, L.}, \bibinfo{author}{Panine, M.},
  \bibinfo{author}{Rampini, A.}, \bibinfo{author}{Ovsjanikov, M.},
  \bibinfo{author}{Bronstein, M.~M.}, \& \bibinfo{author}{Rodol{\`{a}}, E.}
  (\bibinfo{year}{2019}).
\newblock \bibinfo{title}{Isospectralization, or how to hear shape, style, and
  correspondence}.
\newblock In {\it \bibinfo{booktitle}{CVPR}\/}.
\bibitem[{De~Lange et~al.(2021)De~Lange, Aljundi, Masana, Parisot, Jia,
  Leonardis, Slabaugh \& Tuytelaars}]{de2019continual}
\bibinfo{author}{De~Lange, M.}, \bibinfo{author}{Aljundi, R.},
  \bibinfo{author}{Masana, M.}, \bibinfo{author}{Parisot, S.},
  \bibinfo{author}{Jia, X.}, \bibinfo{author}{Leonardis, A.},
  \bibinfo{author}{Slabaugh, G.}, \& \bibinfo{author}{Tuytelaars, T.}
  (\bibinfo{year}{2021}).
\newblock \bibinfo{title}{{A continual learning survey: Defying forgetting in
  classification tasks}}.
\newblock {\it \bibinfo{journal}{IEEE TPAMI}\/}, .
\bibitem[{Douillard et~al.(2020)Douillard, Cord, Ollion, Robert \&
  Valle}]{douillard2020podnet}
\bibinfo{author}{Douillard, A.}, \bibinfo{author}{Cord, M.},
  \bibinfo{author}{Ollion, C.}, \bibinfo{author}{Robert, T.}, \&
  \bibinfo{author}{Valle, E.} (\bibinfo{year}{2020}).
\newblock \bibinfo{title}{{Podnet: Pooled outputs distillation for small-tasks
  incremental learning}}.
\newblock In {\it \bibinfo{booktitle}{ECCV}\/}.
\bibitem[{Farquhar \& Gal(2018)}]{farquhar2018towards}
\bibinfo{author}{Farquhar, S.}, \& \bibinfo{author}{Gal, Y.}
  (\bibinfo{year}{2018}).
\newblock \bibinfo{title}{{Towards Robust Evaluations of Continual Learning}}.
\newblock In {\it \bibinfo{booktitle}{ICML Workshops}\/}.
\bibitem[{HaoChen et~al.(2021)HaoChen, Wei, Gaidon \& Ma}]{haochen2021provable}
\bibinfo{author}{HaoChen, J.~Z.}, \bibinfo{author}{Wei, C.},
  \bibinfo{author}{Gaidon, A.}, \& \bibinfo{author}{Ma, T.}
  (\bibinfo{year}{2021}).
\newblock \bibinfo{title}{Provable guarantees for self-supervised deep learning
  with spectral contrastive loss}.
\newblock In {\it \bibinfo{booktitle}{ANeurIPS}\/}.
\bibitem[{He et~al.(2016)He, Zhang, Ren \& Sun}]{he2016deep}
\bibinfo{author}{He, K.}, \bibinfo{author}{Zhang, X.}, \bibinfo{author}{Ren,
  S.}, \& \bibinfo{author}{Sun, J.} (\bibinfo{year}{2016}).
\newblock \bibinfo{title}{{Deep residual learning for image recognition}}.
\newblock In {\it \bibinfo{booktitle}{CVPR}\/}.
\bibitem[{Jin et~al.(2020)Jin, Loukas \& JaJa}]{coarsening}
\bibinfo{author}{Jin, Y.}, \bibinfo{author}{Loukas, A.}, \&
  \bibinfo{author}{JaJa, J.} (\bibinfo{year}{2020}).
\newblock \bibinfo{title}{Graph coarsening with preserved spectral properties}.
\newblock In {\it \bibinfo{booktitle}{AISTATS}\/}.
\bibitem[{Khosla et~al.(2020)Khosla, Teterwak, Wang, Sarna, Tian, Isola,
  Maschinot, Liu \& Krishnan}]{khosla2020supervised}
\bibinfo{author}{Khosla, P.}, \bibinfo{author}{Teterwak, P.},
  \bibinfo{author}{Wang, C.}, \bibinfo{author}{Sarna, A.},
  \bibinfo{author}{Tian, Y.}, \bibinfo{author}{Isola, P.},
  \bibinfo{author}{Maschinot, A.}, \bibinfo{author}{Liu, C.}, \&
  \bibinfo{author}{Krishnan, D.} (\bibinfo{year}{2020}).
\newblock \bibinfo{title}{{Supervised Contrastive Learning}}.
\newblock In {\it \bibinfo{booktitle}{ANeurIPS}\/}.
\bibitem[{Kirkpatrick et~al.(2017)Kirkpatrick, Pascanu, Rabinowitz, Veness,
  Desjardins, Rusu, Milan, Quan, Ramalho, Grabska-Barwinska
  et~al.}]{kirkpatrick2017overcoming}
\bibinfo{author}{Kirkpatrick, J.}, \bibinfo{author}{Pascanu, R.},
  \bibinfo{author}{Rabinowitz, N.}, \bibinfo{author}{Veness, J.},
  \bibinfo{author}{Desjardins, G.}, \bibinfo{author}{Rusu, A.~A.},
  \bibinfo{author}{Milan, K.}, \bibinfo{author}{Quan, J.},
  \bibinfo{author}{Ramalho, T.}, \bibinfo{author}{Grabska-Barwinska, A.} et~al.
  (\bibinfo{year}{2017}).
\newblock \bibinfo{title}{{Overcoming catastrophic forgetting in neural
  networks}}.
\newblock {\it \bibinfo{journal}{PNAS}\/}, .
\bibitem[{Krizhevsky et~al.(2009)}]{krizhevsky2009learning}
\bibinfo{author}{Krizhevsky, A.} et~al. (\bibinfo{year}{2009}).
\newblock {\it \bibinfo{title}{{Learning multiple layers of features from tiny
  images}}\/}.
\newblock \bibinfo{type}{Technical Report} Citeseer.
\bibitem[{Lassance et~al.(2021)Lassance, Gripon \&
  Ortega}]{lassance2021representing}
\bibinfo{author}{Lassance, C.}, \bibinfo{author}{Gripon, V.}, \&
  \bibinfo{author}{Ortega, A.} (\bibinfo{year}{2021}).
\newblock \bibinfo{title}{{Representing deep neural networks latent space
  geometries with graphs}}.
\newblock {\it \bibinfo{journal}{Algorithms}\/}, .
\bibitem[{Lecun et~al.(1998)Lecun, Bottou, Bengio \& Haffner}]{lecun1998mnist}
\bibinfo{author}{Lecun, Y.}, \bibinfo{author}{Bottou, L.},
  \bibinfo{author}{Bengio, Y.}, \& \bibinfo{author}{Haffner, P.}
  (\bibinfo{year}{1998}).
\newblock \bibinfo{title}{Gradient-based learning applied to document
  recognition}.
\newblock {\it \bibinfo{journal}{Proceedings of the IEEE}\/},  {\it
  \bibinfo{volume}{86}\/}, \bibinfo{pages}{2278--2324}.
  \DOIprefix\doi{10.1109/5.726791}.
\bibitem[{Lee et~al.(2014)Lee, Gharan \& Trevisan}]{trevisan14}
\bibinfo{author}{Lee, J.~R.}, \bibinfo{author}{Gharan, S.~O.}, \&
  \bibinfo{author}{Trevisan, L.} (\bibinfo{year}{2014}).
\newblock \bibinfo{title}{Multiway spectral partitioning and higher-order
  cheeger inequalities}.
\newblock {\it \bibinfo{journal}{J. ACM}\/}, .
\bibitem[{Lopez-Paz \& Ranzato(2017)}]{lopez2017gradient}
\bibinfo{author}{Lopez-Paz, D.}, \& \bibinfo{author}{Ranzato, M.}
  (\bibinfo{year}{2017}).
\newblock \bibinfo{title}{{Gradient episodic memory for continual learning}}.
\newblock In {\it \bibinfo{booktitle}{ANeurIPS}\/}.
\bibitem[{Mai et~al.(2021)Mai, Li, Kim \& Sanner}]{mai2021supervised}
\bibinfo{author}{Mai, Z.}, \bibinfo{author}{Li, R.}, \bibinfo{author}{Kim, H.},
  \& \bibinfo{author}{Sanner, S.} (\bibinfo{year}{2021}).
\newblock \bibinfo{title}{Supervised contrastive replay: Revisiting the nearest
  class mean classifier in online class-incremental continual learning}.
\newblock In {\it \bibinfo{booktitle}{CVPR Workshops}\/} (pp.
  \bibinfo{pages}{3589--3599}).
\bibitem[{Mallya \& Lazebnik(2018)}]{mallya2018packnet}
\bibinfo{author}{Mallya, A.}, \& \bibinfo{author}{Lazebnik, S.}
  (\bibinfo{year}{2018}).
\newblock \bibinfo{title}{{Packnet: Adding multiple tasks to a single network
  by iterative pruning}}.
\newblock In {\it \bibinfo{booktitle}{CVPR}\/}.
\bibitem[{Marin et~al.(2020)Marin, Rampini, Castellani, Rodol{\`{a}},
  Ovsjanikov \& Melzi}]{marin2020instant}
\bibinfo{author}{Marin, R.}, \bibinfo{author}{Rampini, A.},
  \bibinfo{author}{Castellani, U.}, \bibinfo{author}{Rodol{\`{a}}, E.},
  \bibinfo{author}{Ovsjanikov, M.}, \& \bibinfo{author}{Melzi, S.}
  (\bibinfo{year}{2020}).
\newblock \bibinfo{title}{Instant recovery of shape from spectrum via latent
  space connections}.
\newblock In \bibinfo{editor}{V.~Struc}, \& \bibinfo{editor}{F.~G.
  Fern{\'{a}}ndez} (Eds.), {\it \bibinfo{booktitle}{3DV}\/}.
\bibitem[{McCloskey \& Cohen(1989)}]{mccloskey1989catastrophic}
\bibinfo{author}{McCloskey, M.}, \& \bibinfo{author}{Cohen, N.~J.}
  (\bibinfo{year}{1989}).
\newblock \bibinfo{title}{{Catastrophic interference in connectionist networks:
  The sequential learning problem}}.
\newblock {\it \bibinfo{journal}{Psychol. Learn. Motiv.}\/}, .
\bibitem[{Moschella et~al.(2022)Moschella, Melzi, Cosmo, Maggioli, Litany,
  Ovsjanikov, Guibas \& Rodol{\`{a}}}]{moschella2022learning}
\bibinfo{author}{Moschella, L.}, \bibinfo{author}{Melzi, S.},
  \bibinfo{author}{Cosmo, L.}, \bibinfo{author}{Maggioli, F.},
  \bibinfo{author}{Litany, O.}, \bibinfo{author}{Ovsjanikov, M.},
  \bibinfo{author}{Guibas, L.~J.}, \& \bibinfo{author}{Rodol{\`{a}}, E.}
  (\bibinfo{year}{2022}).
\newblock \bibinfo{title}{Learning spectral unions of partial deformable 3d
  shapes}.
\newblock {\it \bibinfo{journal}{Comput. Graph. Forum}\/}, .
\bibitem[{Ovsjanikov et~al.(2012)Ovsjanikov, Ben-Chen, Solomon, Butscher \&
  Guibas}]{ovsjanikov2012functional}
\bibinfo{author}{Ovsjanikov, M.}, \bibinfo{author}{Ben-Chen, M.},
  \bibinfo{author}{Solomon, J.}, \bibinfo{author}{Butscher, A.}, \&
  \bibinfo{author}{Guibas, L.} (\bibinfo{year}{2012}).
\newblock \bibinfo{title}{Functional maps: a flexible representation of maps
  between shapes}.
\newblock {\it \bibinfo{journal}{ACM Transactions on Graphics (ToG)}\/}, .
\bibitem[{Pernici et~al.(2021)Pernici, Bruni, Baecchi, Turchini \&
  Del~Bimbo}]{pernici2021class}
\bibinfo{author}{Pernici, F.}, \bibinfo{author}{Bruni, M.},
  \bibinfo{author}{Baecchi, C.}, \bibinfo{author}{Turchini, F.}, \&
  \bibinfo{author}{Del~Bimbo, A.} (\bibinfo{year}{2021}).
\newblock \bibinfo{title}{{Class-incremental learning with pre-allocated fixed
  classifiers}}.
\newblock In {\it \bibinfo{booktitle}{ICPR}\/}.
\bibitem[{Rampini et~al.(2021)Rampini, Pestarini, Cosmo, Melzi \&
  Rodol{\`{a}}}]{rampini2021universal}
\bibinfo{author}{Rampini, A.}, \bibinfo{author}{Pestarini, F.},
  \bibinfo{author}{Cosmo, L.}, \bibinfo{author}{Melzi, S.}, \&
  \bibinfo{author}{Rodol{\`{a}}, E.} (\bibinfo{year}{2021}).
\newblock \bibinfo{title}{Universal spectral adversarial attacks for deformable
  shapes}.
\newblock In {\it \bibinfo{booktitle}{CVPR}\/}.
\bibitem[{Rebuffi et~al.(2017)Rebuffi, Kolesnikov, Sperl \&
  Lampert}]{rebuffi2017icarl}
\bibinfo{author}{Rebuffi, S.-A.}, \bibinfo{author}{Kolesnikov, A.},
  \bibinfo{author}{Sperl, G.}, \& \bibinfo{author}{Lampert, C.~H.}
  (\bibinfo{year}{2017}).
\newblock \bibinfo{title}{{iCaRL: Incremental classifier and representation
  learning}}.
\newblock In {\it \bibinfo{booktitle}{CVPR}\/}.
\bibitem[{Rodol{\`a} et~al.(2017)Rodol{\`a}, Cosmo, Bronstein, Torsello \&
  Cremers}]{rodola2017partial}
\bibinfo{author}{Rodol{\`a}, E.}, \bibinfo{author}{Cosmo, L.},
  \bibinfo{author}{Bronstein, M.~M.}, \bibinfo{author}{Torsello, A.}, \&
  \bibinfo{author}{Cremers, D.} (\bibinfo{year}{2017}).
\newblock \bibinfo{title}{Partial functional correspondence}.
\newblock In {\it \bibinfo{booktitle}{Comput. Graph. Forum}\/}.
\bibitem[{Shao et~al.(2018)Shao, Kumar \& Fletcher}]{shao2019riemannian}
\bibinfo{author}{Shao, H.}, \bibinfo{author}{Kumar, A.}, \&
  \bibinfo{author}{Fletcher, P.~T.} (\bibinfo{year}{2018}).
\newblock \bibinfo{title}{The riemannian geometry of deep generative models}.
\newblock In {\it \bibinfo{booktitle}{CVPR Workshops}\/}.
\bibitem[{Shi \& Malik(2000)}]{shi2000normalized}
\bibinfo{author}{Shi, J.}, \& \bibinfo{author}{Malik, J.}
  (\bibinfo{year}{2000}).
\newblock \bibinfo{title}{Normalized cuts and image segmentation}.
\newblock {\it \bibinfo{journal}{IEEE TPAMI}\/}, .
\bibitem[{Sinclair \& Jerrum(1989)}]{sinclair1989approximate}
\bibinfo{author}{Sinclair, A.}, \& \bibinfo{author}{Jerrum, M.}
  (\bibinfo{year}{1989}).
\newblock \bibinfo{title}{Approximate counting, uniform generation and rapidly
  mixing markov chains}.
\newblock {\it \bibinfo{journal}{Inf. Comput.}\/}, .
\bibitem[{Tan \& Le(2019)}]{tan2019efficientnet}
\bibinfo{author}{Tan, M.}, \& \bibinfo{author}{Le, Q.} (\bibinfo{year}{2019}).
\newblock \bibinfo{title}{{Efficientnet: Rethinking model scaling for
  convolutional neural networks}}.
\newblock In {\it \bibinfo{booktitle}{ICML}\/}.
\bibitem[{Vaswani et~al.(2017)Vaswani, Shazeer, Parmar, Uszkoreit, Jones,
  Gomez, Kaiser \& Polosukhin}]{vaswani2017attention}
\bibinfo{author}{Vaswani, A.}, \bibinfo{author}{Shazeer, N.},
  \bibinfo{author}{Parmar, N.}, \bibinfo{author}{Uszkoreit, J.},
  \bibinfo{author}{Jones, L.}, \bibinfo{author}{Gomez, A.~N.},
  \bibinfo{author}{Kaiser, L.~u.}, \& \bibinfo{author}{Polosukhin, I.}
  (\bibinfo{year}{2017}).
\newblock \bibinfo{title}{Attention is all you need}.
\newblock In {\it \bibinfo{booktitle}{ANeurIPS}\/}.
\bibitem[{van~de Ven et~al.(2022)van~de Ven, Tuytelaars \&
  Tolias}]{van2019three}
\bibinfo{author}{van~de Ven, G.~M.}, \bibinfo{author}{Tuytelaars, T.}, \&
  \bibinfo{author}{Tolias, A.~S.} (\bibinfo{year}{2022}).
\newblock \bibinfo{title}{{Three types of incremental learning}}.
\newblock {\it \bibinfo{journal}{Nat. Mach. Intell.}\/}, .
\bibitem[{Vinyals et~al.(2016)Vinyals, Blundell, Lillicrap, Wierstra
  et~al.}]{vinyals2016matching}
\bibinfo{author}{Vinyals, O.}, \bibinfo{author}{Blundell, C.},
  \bibinfo{author}{Lillicrap, T.}, \bibinfo{author}{Wierstra, D.} et~al.
  (\bibinfo{year}{2016}).
\newblock \bibinfo{title}{{Matching networks for one shot learning}}.
\newblock In {\it \bibinfo{booktitle}{ANeurIPS}\/}.
\bibitem[{Wang et~al.(2022)Wang, Zhang, Lee, Zhang, Sun, Ren, Su, Perot, Dy \&
  Pfister}]{wang2022learning}
\bibinfo{author}{Wang, Z.}, \bibinfo{author}{Zhang, Z.}, \bibinfo{author}{Lee,
  C.-Y.}, \bibinfo{author}{Zhang, H.}, \bibinfo{author}{Sun, R.},
  \bibinfo{author}{Ren, X.}, \bibinfo{author}{Su, G.}, \bibinfo{author}{Perot,
  V.}, \bibinfo{author}{Dy, J.}, \& \bibinfo{author}{Pfister, T.}
  (\bibinfo{year}{2022}).
\newblock \bibinfo{title}{Learning to prompt for continual learning}.
\newblock In {\it \bibinfo{booktitle}{CVPR}\/}.
\bibitem[{Wu et~al.(2019)Wu, Chen, Wang, Ye, Liu, Guo \& Fu}]{wu2019large}
\bibinfo{author}{Wu, Y.}, \bibinfo{author}{Chen, Y.}, \bibinfo{author}{Wang,
  L.}, \bibinfo{author}{Ye, Y.}, \bibinfo{author}{Liu, Z.},
  \bibinfo{author}{Guo, Y.}, \& \bibinfo{author}{Fu, Y.}
  (\bibinfo{year}{2019}).
\newblock \bibinfo{title}{{Large scale incremental learning}}.
\newblock In {\it \bibinfo{booktitle}{CVPR}\/}.
\bibitem[{Wu et~al.(2018)Wu, Xiong, Yu \& Lin}]{wu2018unsupervised}
\bibinfo{author}{Wu, Z.}, \bibinfo{author}{Xiong, Y.}, \bibinfo{author}{Yu,
  S.~X.}, \& \bibinfo{author}{Lin, D.} (\bibinfo{year}{2018}).
\newblock \bibinfo{title}{Unsupervised feature learning via non-parametric
  instance discrimination}.
\newblock In {\it \bibinfo{booktitle}{CVPR}\/}.

\end{thebibliography}

\clearpage
\setcounter{page}{1}
\onecolumn

\renewcommand{\thetable}{\Alph{table}}
\renewcommand{\theequation}{\Alph{equation}}
\renewcommand{\thefigure}{\Alph{figure}}

\makeatletter
\renewcommand{\maketitle}{
\begin{center}

\pagestyle{empty}
\phantom{.}  
\vspace{2cm}
{\LARGE \bf \@title\par}
\vspace{2.5cm}
\end{center}
}\makeatother

\title{\\Supplemental Material} 

\appendix
\maketitle

\section{Details of \methnam}
\label{appendix:method-details}
In Alg.~\ref{alg:casper} the reader can find the steps to compute \methnam's loss. In detail:
\begin{itemize}
    \item In line 1: the $BalancedSampling$ function extract $b$ examples from the buffer belonging to $p$ different classes uniformly.
    \item In line 4: to compute the $k$-NN a hyperparameter $k$ is needed; after finding the $k$ nearest neighbors of every point, the adjacency matrix is then symmetrized; therefore, the number of neighbors at the end may become higher than $k$; 
\end{itemize}

\begin{algorithm}[H]
\caption{\methnam Loss Computation}\label{alg:casper}
\textbf{Input} Memory buffer $B$ of saved samples, with size $m$; batch-size $b$; \\ 
\hspace*{2.5em} hyperparameter of number of classes to sample $p$\\
\vspace{-1.3em}
\begin{algorithmic}[1]
\STATE $\boldsymbol{x}^b \gets \operatorname{BalancedSampling}(B, p)$ 
\COMMENT{Data extraction from buffer}
\STATE $\boldsymbol{z}^b \gets F_{\boldsymbol{\theta}^f}(\boldsymbol{x}^b)$ \COMMENT{Get features}
\STATE $\boldsymbol{\mathcal{D}}=\{D_{i,j} = \lVert z^b_i - z^b_j \rVert \: \text{with} \: i, j = 1, ..., m\}$ 
\COMMENT{Calculate distance matrix}
\STATE $\boldsymbol{A} \gets \operatorname{k-NN}(\mathcal{D})$ 
\COMMENT{Retrieve adjacency matrix}
\STATE $\boldsymbol{D} \gets \operatorname{diag}(\sum_i^{m} a_{1,i}, \sum_i^{m} a_{2,i}, ..., \sum_i^{m} a_{b,i})$
\COMMENT{Calculate degree matrix}
\STATE $\boldsymbol{L} \gets \boldsymbol{I} - \boldsymbol{D}^{-1/2} \boldsymbol{A} \boldsymbol{D}^{-1/2}$ 
\COMMENT{Compute normalized laplacian}
\STATE $ \boldsymbol{\lambda} \gets \operatorname{Eigenvalues}(\boldsymbol{L}) $\: 
\COMMENT{Retrieve eigenvalues}
\STATE $\ell_{\text{CaSpeR}} \gets -\lambda_{p+1} + \sum_{j=1}^p \lambda_j$  
\COMMENT{Casper loss}
\end{algorithmic}
\textbf{Output} $\ell_{\text{CaSpeR}}$
\end{algorithm}

\section{Adopded Baselines}
Here we briefly explain the rehearsal methods utilized in our experiments:
\begin{itemize}
    \item \textbf{Experience Replay with Asymmetric Cross-Entropy (ER-ACE)}~\citep{caccia2022new}: starting from classic Experience Replay, the authors obtain a significant performance gain by freezing the previous task heads of the classifier while computing the loss on the streaming data.
    \item \textbf{Incremental Classifier and Representation Learning (iCaRL)}~\citep{rebuffi2017icarl}: 
    this method seeks to learn the best representation of data that fits a nearest-neighbor classifier w.r.t.\ class prototypes stored in the buffer.
    \item \textbf{Dark Experience Replay (DER++)}~\citep{buzzega2020dark}: another variant of ER, which combines the standard classification replay with a distillation loss.
    \item \textbf{eXtended-DER (X-DER)}~\citep{boschini2022class}: a method which improves \dpp by addressing its shortcomings and focusing on organically accommodating future knowledge; specifically, we use the more efficient baseline based on a Regular Polytope Classifier~\citep{pernici2021class}.
    \item \textbf{Pooled Outputs Distillation Network (PODNet)}~\citep{douillard2020podnet}: the authors extend iCarl's classification method: their model  learns multiple representations for each class and adopts two additional distillation losses.
\end{itemize}

\section{Additional training details}
For our experiments in Sec.~\ref{sec:exps}, we adopt a batch size of 64 examples. The same number is employed when sampling data from the memory buffer, always extracting 64 data points. For the experiments on Split CIFAR-10 and Split CIFAR-100, the training procedure of each task lasts 20 epochs per task, without a learning rate scheduler. For Split \miniimagenet we train the model for 50 epochs per task, with a learning rate decay of 0.1 applied at epochs 35 and 45. We do not change these choices within each method to grant fairness when comparing them.

\section{Different Incremental Benchmarks}
\subsection{Task Incremental Learning}
In Task-Incremental Learning (Task-IL), the task information is given both during training and evaluation; while being considered easier than Class-IL, it may be especially relevant for the quantification of forgetting, as it is unaffected by data imbalance biases~\citep{wu2019large, boschini2022class}.
Results in Tab.~\ref{tab:til} suggest that \methnam allows the model to better learn and consolidate each task individually (Task-IL). 

\begin{table*}[t]
\small
\caption{Task-IL results -- $\bar{A}_F$ ($\bar{F}^*_F$) -- for SOTA rehearsal CL methods, with and without \methnam.} \label{tab:til}

\begin{center}
\rowcolors{5}{lightgray}{}
\begin{tabular}{l@{\hskip 0.5cm}cc@{\hskip 0.5cm}cc}
\hline
\textbf{Task-IL} & \multicolumn{2}{c}{\textbf{Split CIFAR-100}} & \multicolumn{2}{c}{\textbf{Split \miniimagenet}}\\
\hline
Joint (UB)                          & \multicolumn{2}{c}{\resultAF{88.81}{0.84}{-}{-}}                & \multicolumn{2}{c}{\resultAF{87.39}{0.46}{-}{-}} \\
Finetune (LB)                       & \multicolumn{2}{c}{\resultAF{30.10}{5.61}{43.88}{62.84}}            & \multicolumn{2}{c}{\resultAF{24.05}{0.33}{45.71}{67.37}} \\
\hline
\textbf{Buffer Size} & 500 & 2000 & 2000 & 5000\\
\hline 
ER-ACE                              & \resultAF{73.86}{0.93}{ 9.76}{10.73} & \resultAF{80.69}{1.07}{ 4.40}{ 5.37}  & \resultAF{69.05}{0.48}{}{13.72} & \resultAF{72.78}{0.55}{}{8.93} \\
\hspace{.3em} + \textbf{\methnam}   & 
\resultAFG{75.14}{0.42}{ 9.69}{ 4.91}{1.28} & 
\resultAFG{81.57}{0.54}{}{ 4.93}{0.88}  & 
\resultAFG{69.59}{1.17}{10.18}{13.05}{0.54} & 
\resultAFG{74.14}{0.18}{}{8.12}{1.36} \\
iCaRL                               & \resultAF{78.38}{1.02}{ 4.15}{ 5.38} & \resultAF{78.47}{1.08}{ 3.72}{ 4.91}  & \resultAF{70.35}{0.49}{ 2.21}{ 3.92} & \resultAF{70.44}{0.36}{}{2.68} \\
\hspace{.3em} + \textbf{\methnam}   & \resultAFG{79.31}{0.67}{}{4.61}{0.93} & \resultAFG{79.43}{0.62}{ 3.07}{ 3.41}{0.96}  & \resultAFG{71.19}{0.67}{ 2.71}{ 3.67}{0.84} & \resultAFG{71.93}{0.35}{2.67}{3.65}{1.49} \\
\dpp                                & \resultAF{70.55}{1.72}{}{11.12} & \resultAF{78.60}{1.90}{}{ 5.96}  & \resultAF{69.78}{0.45}{}{13.37} & \resultAF{73.81}{0.32}{6.79}{8.59} \\
\hspace{.3em} + \textbf{\methnam}   & \resultAFG{73.25}{1.63}{}{9.49}{2.70} & \resultAFG{80.78}{0.42}{ 3.43}{ 3.04}{2.18}  & \resultAFG{70.97}{0.57}{}{11.75}{1.19} & \resultAFG{75.18}{0.30}{}{7.93}{1.37} \\
X-DER                           & \resultAF{77.28}{0.27}{ 2.81}{ 2.43} & \resultAF{82.55}{2.98}{ 0.65}{ 0.92}  & \resultAF{74.32}{0.33}{ 3.81}{ 4.95} & \resultAF{77.70}{0.56}{2.93}{3.71} \\
\hspace{.3em} + \textbf{\methnam}   & \resultAFG{78.26}{1.82}{ 6.79}{ 5.47}{0.98} & \resultAFG{83.77}{2.09}{ 0.56}{ 0.27}{1.22}  & \resultAFG{75.99}{0.22}{ 3.03}{ 3.88}{1.67} & \resultAFG{78.71}{0.53}{1.92}{2.32}{1.01} \\
PODNet                              & \resultAF{67.37}{1.55}{}{19.76} & \resultAF{69.63}{2.48}{}{15.16}  & \resultAF{60.60}{1.94}{}{14.00} & \resultAF{66.15}{2.81}{}{10.71} \\
\hspace{.3em} + \textbf{\methnam}   & \resultAFG{70.81}{1.71}{}{15.26}{3.44} & \resultAFG{71.90}{0.65}{10.77}{11.32}{2.27}  & \resultAFG{64.84}{1.80}{}{10.01}{4.24} & \resultAFG{70.85}{0.62}{}{7.99}{4.70} \\
\hline
\end{tabular}
\end{center}
\end{table*}

\subsection{Domain Incremental Learning}
In Domain Incremental Learning (Domain-IL) scenarios, all classes are available to the learner from the initial task. Differently from Class-IL and Task-IL, the label target distribution remains the same, but the associated input distribution undergoes a shift at each task. In \textbf{Rotated MNIST}~\citep{lopez2017gradient}, the learner needs to classify all MNIST~\citep{lecun1998mnist} digits for 20 subsequent tasks, in which images are rotated by a random angle in the interval $[0, \pi[$ (different for each task). 

We evaluate \methnam following the experimental setup of \citep{buzzega2020dark}, excluding the approaches that are not compatible with Domain-IL. Results in Tab.~\ref{tab:domain} show that the benefits of \methnam can lead to a performance improvement even in this setting.

\section{Additional Experiments}
As highlighted in literature \citep{aljundi2019gradient, farquhar2018towards, buzzega2020dark}, comparing results from different CL works is non-trivial due to the effect of even subtler changes in the experimental setting (which should be always shared across all compared methods for fairness). For this reason, we did not base our experiments on pre-existing results, but rather ran all our experiments from scratch applying the same conditions to all evaluated methods. 

Here we provide more experiments with different training configurations to further verify the effectiveness of \methnam.

\textbf{Longer training-phase}.~We replicate the setting of \citep{boschini2022class} for Split CIFAR-100: batch size set at $32$, $50$ epochs per task, and a learning rate decay of 0.1 applied at epochs 35 and 45.
Results, shown in Tab.~\ref{tab:derset}, demonstrate that \methnam can successfully improve different baselines regardless of the experimental setting.

\textbf{Different batch-sizes}.~In Tab.~\ref{tab:batchsizes}, we report additional results for Split CIFAR-100 (Class-IL) revealing that the effectiveness of \methnam is not compromised by the reduced batch size (w.r.t.\ 64 as used in the paper).

\begin{table}[t]
    \centering
  \begin{minipage}{.45\linewidth}
    \centering
    \caption{Class-IL $\bar{A}_F$ values on Split CIFAR-100, with 50 epochs per task and batch size 32 (as for \citep{boschini2022class}).}
    \label{tab:derset}
    \begin{center}
\rowcolors{3}{}{lightgray}
\begin{tabular}{lcc}
\hline
\textbf{Class-IL} & \multicolumn{2}{c}{\textbf{Split CIFAR-100}} \\
\hline
\textbf{Buffer Size} & 500 & 2000 \\ 
\hline

ER-ACE                               & $37.12$ & $49.20$ \\ 
\hspace{.3em} + \textbf{\methnam}    & $38.40$ & $50.38$ \\ 
iCaRL                                & $47.47$ & $52.68$ \\ 
\hspace{.3em} + \textbf{\methnam}    & $48.81$ & $54.44$ \\ 
\dpp                                 & $36.28$ & $50.66$ \\ 
\hspace{.3em} + \textbf{\methnam}    & $40.44$ & $52.22$ \\ 
X-DER                                & $46.96$ & $55.85$ \\ 
\hspace{.3em} + \textbf{\methnam}    & $48.43$ & $56.40$ \\ 
PODNet                               & $36.44$ & $43.97$ \\ 
\hspace{.3em} + \textbf{\methnam}    & $40.63$ & $46.80$ \\ 

\hline
\end{tabular}
\end{center}
  \end{minipage}
  \hspace{.05\linewidth}
  \begin{minipage}{.45\linewidth}
    \centering
    \caption{Class-IL results -- $\bar{A}_F$ -- in Split CIFAR-100 for different batch size configurations (20 epochs per task). Buffer size $500$.} 
    \label{tab:batchsizes}
    
\begin{center}
\rowcolors{3}{}{lightgray}
\begin{tabular}{lcc}
\hline
\textbf{Class-IL} & \multicolumn{2}{c}{\textbf{Split CIFAR-100}} \\
\hline
\textbf{Batch Size} & 16 & 32\\ 
\hline

ER-ACE                               & $35.70$ & $36.47$ \\ 
\hspace{.3em} + \textbf{\methnam}    & $36.96$ & $38.63$ \\ 
iCaRL                                & $43.12$ & $43.99$ \\ 
\hspace{.3em} + \textbf{\methnam}    & $44.22$ & $45.20$ \\ 
\dpp                                 & $28.91$ & $31.14$ \\ 
\hspace{.3em} + \textbf{\methnam}    & $31.93$ & $33.20$ \\ 
X-DER                                & $33.02$ & $39.19$ \\ 
\hspace{.3em} + \textbf{\methnam}    & $35.42$ & $40.46$ \\ 
PODNet                               & $34.29$ & $35.22$ \\ 
\hspace{.3em} + \textbf{\methnam}    & $36.15$ & $37.02$ \\ 

\hline
\end{tabular}
\end{center}
  \end{minipage}
\end{table}

\begin{table}[t]
    \centering
  \begin{minipage}{.45\linewidth}
    \centering
    \caption{$\bar{A}_F$ values on Rotated MNIST.}
    \label{tab:domain}
    \begin{center}
\setlength{\tabcolsep}{2pt}
\rowcolors{3}{lightgray}{}
\begin{tabular}{lcc}
\hline

\textbf{Domain-IL} & \multicolumn{2}{c}{Rotated MNIST}\\
\hline
Joint (UB)       & \multicolumn{2}{c}{95.76} \\
Finetune (LB)    & \multicolumn{2}{c}{67.66} \\
\hline
\textbf{Buffer Size} & $200$ & $500$\\
\hline

ER-ACE             & $83.45$ & $86.86$ \\
\hspace{.3em} + \textbf{\methnam} & $86.19$ & $88.11$ \\

\dpp               & $89.91$ & $92.00$ \\
\hspace{.3em} + \textbf{\methnam} & $90.96$ & $93.21$ \\

\hline
\end{tabular}
\end{center}
  \end{minipage}
  \hspace{.05\linewidth}
  \begin{minipage}{.45\linewidth}
    \centering
    \caption{Class-IL $\bar{A}_F$ values on Split CIFAR-100, with reduced amount of annotations (CSSL). Buffer size $2000$.}
    \label{tab:cssl}
    \begin{center}
\rowcolors{3}{lightgray}{}
\begin{tabular}{lcc}
\hline
\multicolumn{3}{c}{\textbf{CSSL}} \\ 
Labels \%  & $0.8\%$ & $5\%$ \\ 
\hline
ER-ACE & $8.46$ & $11.87$ \\ 
\hspace{.3em} + \textbf{\methnam} & $8.55$ & $14.16$ \\
PsER-ACE & $2.31$ & $16.35$ \\ 
\hspace{.3em} + \textbf{\methnam} & $9.69$ & $17.42$ \\
CCIC & $11.5^\dagger$ & $19.5^\dagger$ \\ 
\hspace{.3em} + \textbf{\methnam} & $12.22$ & $20.32$ \\
\hline
\end{tabular}
\end{center}
  \end{minipage}
\end{table}

\section{Continual Semi-supervised Learning}
In a supervised CL setting, we apply \methnam to buffer data points, thus encouraging the separation of all previously encountered classes in the latent space. However, our proposed approach does not have strict supervision requirements, as it does not need the labels attached to each node in the LGG, but rather just the total amount of classes $g$ that must be clustered (Eq.~\ref{eqn:casperloss}).
With the assumption that classes are equally distributed in our data, the \textit{balanced sampling} (explained in Sec. \ref{appendix:method-details}) can be approximated with random sampling from the memory buffer. Therefore, \methnam will not rely on the availability of annotations for each example, allowing it to be applied to semi-supervised scenarios to provide better accuracy and easier convergence.

\citep{boschini2022continual} propose Continual Semi-Supervised Learning (CSSL), a new CL experimental benchmark where only a fraction of the examples on the input stream are associated with an annotation.
In Tab.~\ref{tab:cssl}, we report the results of an experiment on Split CIFAR-100 in the CSSL setting with only $0.8\%$ or $5\%$ annotated labels. Typical CL methods operating in this scenario are forced to discard a consistent amount of data (ER-ACE), leading to majorly reduced performance w.r.t.\ the fully-supervised case, or to use the in-training model to annotate unlabeled samples (\textit{pseudo-labeling}, PsER-ACE), but might backfire if the provided supervision does not suffice for the learner to produce reliable responses (as is the case with $0.8\%$ labels).

To allow for the exploitation of unlabeled exemplars, we also apply \methnam on data points from the input stream, by taking $k$ equal to the number of classes in a given task. We show that this leads to an overall improvement of the tested models and -- particularly -- counteracts the failure case where PseudoER-ACE is applied on top of a few annotated data. This indicates that \methnam manages to limit the impact of the noisy labels produced by \textit{pseudo-labeling}. 

While our proposed regularizer can moderately operate without full supervision, we remark that it still depends on the availability of supervised training signals. The applicability of geometric-based constraints to unsupervised or self-supervised CL scenarios is still a work in progress.

\section{An analysis of training time}
\label{sec:wallclock}
\begin{figure*}[t]
    \centering
    \includegraphics[width=.65\linewidth]{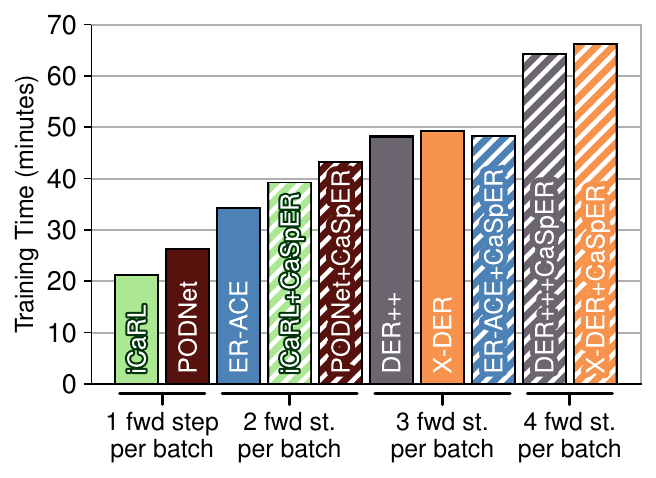}
    \caption{Wallclock training time for the approaches evaluated in Sec.~\ref{sec:exps} benchmarked on an identical hardware setup (GPU NVIDIA V100). Training time grows linearly w.r.t.\ the number of per-batch forward steps.}
    \label{fig:times}
\end{figure*}

As outlined in Sec.~\ref{sec:method_continual}, \methnam involves non-negligible computations that might seem to have a significant impact on the overall training time. In this section, we shed light on the matter of the efficiency of our approximated procedure by comparing the wall-clock time between models with and without \methnam on identical run conditions.

In Fig.~\ref{fig:times}, we report the overall training time for the approaches evaluated in Sec.~\ref{sec:exps}, specifically on Split CIFAR-100, benchmarked on an identical hardware setup. We observe that the overhead of \methnam is fundamentally dominated by the time it takes to perform an additional forward step of the corresponding batch through the model. For X-DER, which already performs 3 forward steps, we register a 33\% increase in compute time; for ER-ACE, which performs 2, we get a 43\% increase, and so on. We believe that such an overhead is comparable to the ones of other CL solutions (for instance, meaning that ER-ACE+\methnam ends up having roughly the same compute time as X-DER and DER++).

\section{Hyperparameters selection}
The optimal configuration of every baseline was found via grid search.\\
While \methnam adds three hyperparameters to each method, the degrees of freedom to grid them are actually two: the regularization weight $\rho$, which is independent; $p$, the number of classes to sample from the buffer, and $k$, number of nearest-neighbor for the adjacency matrix, are linked. Specifically, their product should be equal (or slightly lower) than the number of extracted samples, which in our experiments is the same as the batch size.

\subsection{Split CIFAR-10}%
%
%
\noindent\begin{small}\textbf{No buffer}\end{small} \\
\textbf{Finetune}: $lr$:0.03 \\
\textbf{Joint}: $lr$:0.1 \\

\noindent\begin{small}\textbf{Buffer size: 500}\end{small}\\
\textbf{ER-ACE}: $lr$:0.1\\
\textbf{ER-ACE + \methnam}: $lr$:0.1; ~$p$:8; ~$k$:8; ~$\rho$:0.01\\
\textbf{iCaRL}: $lr$:0.01; ~$wd$:1e-05\\
\textbf{iCaRL + \methnam}: $lr$:0.01; ~$wd$:1e-05; ~$p$:2; ~$k$:20; ~$\rho$:0.001\\
\textbf{DER++}: $lr$:0.03; $\alpha$:0.2; $\beta$:0.5\\
\textbf{DER++ + \methnam}: $lr$:0.03; $\alpha$:0.2; $\beta$:0.5; ~$p$:8; ~$k$:8; ~$\rho$:0.001\\
\textbf{X-DER}: $lr$:0.03; $\alpha$:0.2; $\beta$:0.9; $\gamma$:0.85; $\eta$:0.001; $m$:0.3\\
\textbf{X-DER + \methnam}: $lr$:0.03; $\alpha$:0.2; $\beta$:0.9; $\gamma$:0.85; $\eta$:0.001; $m$:0.3; ~$p$:6; ~$k$:10; ~$\rho$:0.01\\
\textbf{PODNet}: $lr$:0.03; $wd$:5e-05; $\lambda_c$:0.05; $\lambda_f$:0.01; $\delta$:0.6; $\eta$:1; ~$k^{\text{PODNet}}$:10; ~$scaling$:3\\
\textbf{PODNet + \methnam}: $lr$:0.03; $wd$:5e-05; $\lambda_c$:0.05; $\lambda_f$:0.01; $\delta$:0.6; $\eta$:1; ~$k^{\text{PODNet}}$:10; ~$scaling$:3; ~$p$:6; ~$k$:10; ~$\rho$:1e-4\\

\noindent\begin{small}\textbf{Buffer size: 1000}\end{small}\\
\textbf{ER-ACE}: $lr$:0.1\\
\textbf{ER-ACE + \methnam}: $lr$:0.1; ~$p$:8; ~$k$:8; ~$\rho$:0.01\\
\textbf{iCaRL}: $lr$:0.01; ~$wd$:1e-05\\
\textbf{iCaRL + \methnam}: $lr$:0.01; ~$wd$:1e-05; ~$p$:8; ~$k$:8; ~$\rho$:0.001\\
\textbf{DER++}: $lr$:0.03; $\alpha$:0.2; $\beta$:0.5\\
\textbf{DER++ + \methnam}: $lr$:0.03; $\alpha$:0.2; $\beta$:0.5; ~$p$:10; ~$k$:6; ~$\rho$:0.001\\
\textbf{X-DER}: $lr$:0.03; $\alpha$:0.2; $\beta$:0.9; $\gamma$:0.85; $\eta$:0.001; $m$:0.3\\
\textbf{X-DER + \methnam}: $lr$:0.03; $\alpha$:0.2; $\beta$:0.9; $\gamma$:0.85; $\eta$:0.001; $m$:0.3; ~$p$:10; ~$k$:6; ~$\rho$:0.01\\
\textbf{PODNet}: $lr$:0.03; $wd$:5e-05; $\lambda_c$:0.05; $\lambda_f$:0.01; $\delta$:0.6; $\eta$:1; ~$k^{\text{PODNet}}$:10; ~$scaling$:3\\
\textbf{PODNet + \methnam}: $lr$:0.03; $wd$:5e-05; $\lambda_c$:0.05; $\lambda_f$:0.01; $\delta$:0.6; $\eta$:1; ~$k^{\text{PODNet}}$:10; ~$scaling$:3; ~$p$:6; ~$k$:10; ~$\rho$:1e-4\\

\subsection{Split CIFAR-100}%
%
%
\noindent\begin{small}\textbf{No buffer}\end{small}\\
\textbf{Finetune}: $lr$:0.1\\
\textbf{Joint}: $lr$:0.1\\

\noindent\begin{small}\textbf{Buffer size: 500}\end{small}\\
\textbf{ER-ACE}: $lr$:0.1\\
\textbf{ER-ACE + \methnam}: $lr$:0.1; ~$p$:16; ~$k$:4; ~$\rho$:0.001\\
\textbf{iCaRL}: $lr$:0.3; ~$wd$:1e-05\\
\textbf{iCaRL + \methnam}: $lr$:0.3; ~$wd$:1e-05; ~$p$:13; ~$k$:5; ~$\rho$:0.001\\
\textbf{DER++}: $lr$:0.1; $\alpha$:0.5; $\beta$:0.1\\
\textbf{DER++ + \methnam}: $lr$:0.1; $\alpha$:0.5; $\beta$:0.1; ~$p$:16; ~$k$:4; ~$\rho$:0.01\\
\textbf{X-DER}: $lr$:0.1; $\alpha$:0.5; $\beta$:0.1; $\gamma$:0.85; $\eta$:0.001; $m$:0.3\\
\textbf{X-DER + \methnam}: $lr$:0.1; $\alpha$:0.5; $\beta$:0.1; $\gamma$:0.85; $\eta$:0.001; $m$:0.3; ~$p$:16; ~$k$:4; ~$\rho$:0.001\\
\textbf{PODNet}: $lr$:0.1; $wd$:5e-05; $\lambda_c$:0.05; $\lambda_f$:0.01; $\delta$:0.6; $\eta$:1; ~$k^{\text{PODNet}}$:10; ~$scaling$:3\\
\textbf{PODNet + \methnam}: $lr$:0.1; $wd$:5e-05; $\lambda_c$:0.05; $\lambda_f$:0.01; $\delta$:0.6; $\eta$:1; ~$k^{\text{PODNet}}$:10; ~$scaling$:3; ~$p$:16; ~$k$:4; ~$\rho$:0.01\\

\noindent\begin{small}\textbf{Buffer size: 2000}\end{small}\\
\textbf{ER-ACE}: $lr$:0.1\\
\textbf{ER-ACE + \methnam}: $lr$:0.1; ~$p$:10; ~$k$:6; ~$\rho$:0.01\\
\textbf{iCaRL}: $lr$:0.3; ~$wd$:1e-05\\
\textbf{iCaRL + \methnam}: $lr$:0.3; ~$wd$:1e-05; ~$p$:8; ~$k$:8; ~$\rho$:0.001\\
\textbf{DER++}: $lr$:0.1; $\alpha$:0.5; $\beta$:0.1\\
\textbf{DER++ + \methnam}: $lr$:0.1; $\alpha$:0.5; $\beta$:0.1; ~$p$:12; ~$k$:5; ~$\rho$:0.001\\
\textbf{X-DER}: $lr$:0.1; $\alpha$:0.5; $\beta$:0.1; $\gamma$:0.85; $\eta$:0.001; $m$:0.3\\
\textbf{X-DER + \methnam}: $lr$:0.1; $\alpha$:0.5; $\beta$:0.1; $\gamma$:0.85; $\eta$:0.001; $m$:0.3; ~$p$:8; ~$k$:8; ~$\rho$:0.001\\
\textbf{PODNet}: $lr$:0.1; $wd$:5e-05; $\lambda_c$:0.05; $\lambda_f$:0.01; $\delta$:0.6; $\eta$:1; ~$k^{\text{PODNet}}$:10; ~$scaling$:3\\
\textbf{PODNet + \methnam}: $lr$:0.1; $wd$:5e-05; $\lambda_c$:0.05; $\lambda_f$:0.01; $\delta$:0.6; $\eta$:1; ~$k^{\text{PODNet}}$:10; ~$scaling$:3; ~$p$:10; ~$k$:6; ~$\rho$:0.01\\

%
%
\subsection{Split.\ \textbf{\textit{mini}}ImageNet}%
%
%
\noindent\begin{small}\textbf{No buffer}\end{small}\\
\textbf{Finetune}: $lr$:0.3\\
\textbf{Joint}: $lr$:0.1\\

\noindent\begin{small}\textbf{Buffer size: 2000}\end{small}\\
\textbf{ER-ACE}: $lr$:0.01\\
\textbf{ER-ACE + \methnam}: $lr$:0.01; ~$p$:5; ~$k$:10; ~$\rho$:0.01\\
\textbf{iCaRL}: $lr$:0.1; ~$wd$:1e-05\\
\textbf{iCaRL + \methnam}: $lr$:0.1; ~$wd$:1e-05; ~$p$:5; ~$k$:12; ~$\rho$:1e-04\\
\textbf{DER++}: $lr$:0.01; $\alpha$:0.3; $\beta$:0.8\\
\textbf{DER++ + \methnam}: $lr$:0.01; $\alpha$:0.3; $\beta$:0.8; ~$p$:16; ~$k$:4; ~$\rho$:0.001\\
\textbf{X-DER}: $lr$:0.01; $\alpha$:0.3; $\beta$:0.8; $\gamma$:0.85; $\eta$:0.01; $m$:0.3\\
\textbf{X-DER + \methnam}: $lr$:0.01; $\alpha$:0.3; $\beta$:0.8; $\gamma$:0.85; $\eta$:0.01; $m$:0.3; ~$p$:16; ~$k$:4; ~$\rho$:1e-04\\
\textbf{PODNet}: $lr$:0.01; $wd$:5e-05; $\lambda_c$:5e-04; $\lambda_f$:3e-04; $\delta$:0.6; $\eta$:1; ~$k^{\text{PODNet}}$:10; ~$scaling$:3\\
\textbf{PODNet + \methnam}: $lr$:0.01; $wd$:5e-05; $\lambda_c$:5e-04; $\lambda_f$:3e-04; $\delta$:0.6; $\eta$:1; ~$k^{\text{PODNet}}$:10; ~$scaling$:3; ~$p$:8; ~$k$:8; ~$\rho$:0.001\\

\noindent\begin{small}\textbf{Buffer size: 5000}\end{small}\\
\textbf{ER-ACE}: $lr$:0.01\\
\textbf{ER-ACE + \methnam}: $lr$:0.01; ~$p$:10; ~$k$:6; ~$\rho$:0.01\\
\textbf{iCaRL}: $lr$:0.1; ~$wd$:1e-05\\
\textbf{iCaRL + \methnam}: $lr$:0.1; ~$wd$:1e-05; ~$p$:5; ~$k$:10; ~$\rho$:1e-04\\
\textbf{DER++}: $lr$:0.01; $\alpha$:0.3; $\beta$:0.8\\
\textbf{DER++ + \methnam}: $lr$:0.01; $\alpha$:0.3; $\beta$:0.8; ~$p$:20; ~$k$:3; ~$\rho$:1e-04\\
\textbf{X-DER}: $lr$:0.01; $\alpha$:0.3; $\beta$:0.8; $\gamma$:0.85; $\eta$:0.01; $m$:0.3\\
\textbf{X-DER + \methnam}: $lr$:0.01; $\alpha$:0.3; $\beta$:0.8; $\gamma$:0.85; $\eta$:0.01; $m$:0.3; ~$p$:16; ~$k$:4; ~$\rho$:1e-04\\
\textbf{PODNet}: $lr$:0.01; $wd$:5e-05; $\lambda_c$:5e-04; $\lambda_f$:3e-04; $\delta$:0.6; $\eta$:1; ~$k^{\text{PODNet}}$:10; ~$scaling$:3\\
\textbf{PODNet + \methnam}: $lr$:0.01; $wd$:5e-05; $\lambda_c$:5e-04; $\lambda_f$:3e-04; $\delta$:0.6; $\eta$:1; ~$k^{\text{PODNet}}$:10; ~$scaling$:3; ~$p$:20; ~$k$:3; ~$\rho$:0.01\\
%
%

\end{document}